\documentclass[preprint,12pt,numerical]{elsarticle}

\usepackage{siunitx}
\usepackage[inline]{enumitem}
\usepackage{graphicx}
\usepackage{float}
\usepackage{amssymb}
\usepackage{amsmath}
\DeclareMathOperator*{\argmax}{arg\,max}

\usepackage[labelformat=simple]{subcaption}

\usepackage[table]{xcolor}
\colorlet{lightgray}{lightgray!30}
\usepackage{natbib}
\usepackage{transparent}
\usepackage{booktabs}
\usepackage{appendix}
\usepackage{longtable}
\usepackage{lscape} %or pdflscape for nicer pdf rendering. let's see what EM accepts.
\usepackage{hyperref}

\def\photo{\mathbf{P}}

\def\height{H}
\def\width{W}

\begin{document}

\begin{frontmatter}

\title{Deep Learning-based Multi Project InP Wafer Simulation for Unsupervised Surface Defect Detection}

\author[1]{Emílio Dolgener Cantú}
\author[2]{Rolf Klemens Wittmann}
\author[2]{Oliver Abdeen}
\author[2]{Patrick Wagner}
\author[1,3,4]{Wojciech Samek}
\author[2]{Moritz Baier}
\author[1,5]{Sebastian Lapuschkin\corref{cor1}}
\cortext[cor1]{Corresponding author.}\ead{sebastian.lapuschkin@hhi.fraunhofer.de}

\affiliation[1]{%
    organization={Department of Artificial Intelligence, Fraunhofer Heinrich Hertz Institute},
    addressline={Einsteinufer 37}, % Street address
    city={Berlin}, % City
    postcode={10587}, % Postcode
    country={Germany}}
\affiliation[2]{%
    organization={Department of Photonic Components, Fraunhofer Heinrich Hertz Institute},
    addressline={Einsteinufer 37}, % Street address
    city={Berlin}, % City
    postcode={10587}, % Postcode
    country={Germany}}
\affiliation[3]{%
    organization={Department of Electrical Engineering and Computer Science, Technische Universität Berlin},
    addressline={Straße des 17. Juni 135}, % Street address
    city={Berlin}, % City
    postcode={10623}, % Postcode
    country={Germany}}
\affiliation[4]{%
    organization={Berlin Institute for the Foundations of Learning and Data},
    addressline={Ernst-Reuter Platz 7}, % Street address
    city={Berlin}, % City
    postcode={10587}, % Postcode
    country={Germany}}
\affiliation[5]{%
    organization={Centre of eXplainable Artificial Intelligence, Technological University Dublin},
    addressline={Park House Grangegorman, 191 North Circular Road},
    city={Dublin}, % City
    postcode={D07 H6K8}, % Postcode
    country={Ireland}}

\begin{abstract}
Quality management in semiconductor manufacturing often relies on template matching with known golden standards.
For Indium-Phosphide (InP) multi-project wafer manufacturing, low production scale and high design variability lead to such golden standards being typically unavailable.
Defect detection, in turn, is manual and labor-intensive.
This work addresses this challenge by proposing a methodology to generate a synthetic golden standard using Deep Neural Networks, trained to simulate photo-realistic InP wafer images from CAD data.
We evaluate various training objectives and assess the quality of the simulated images on both synthetic data and InP wafer photographs.
Our deep-learning-based method outperforms a baseline decision-tree-based approach, enabling the use of a 'simulated golden die' from CAD plans in any user-defined region of a wafer for more efficient defect detection.
We apply our method to a template matching procedure, to demonstrate its practical utility in surface defect detection.
\end{abstract}

\begin{keyword}
visual inspection, defect detection, machine learning, neural networks, golden standard simulation, InP, photonics, MPW
\end{keyword}

\end{frontmatter}

\section{Introduction}

Photonic wafers are manufactured through a complex combination of epitaxial crystal growth~\cite{kum2019epitaxial},
etching, deposition and lithography processes, either on silicon (Si) or indium-phosphide (InP) substrates.
InP photonic integration is desirable because it allows for the integration of lasers and optical amplifiers, smaller scales and operation in higher frequency (to the THz range)~\cite{liehr2020foundry}.
InP wafer manufacturing follows the developments already achieved in silicon-based integration~\cite{kish2017system}, being however still more expensive and prone to manufacturing defects~\cite{soares_inp-based_2019}.
The advent of multi-project wafer (MPW) runs enables the integration of multiple distinct designs on a single wafer, with each design consisting of standardized, modular components. This approach makes small-scale experimental production increasingly viable by reducing per-stakeholder costs through effective cost distribution~\cite{smit2019past, smit2014introduction}.
The number of identical finished dies is much lower than in traditional industrial manufacturing, as it is common for a given prototypical MPW design to be produced only once.
This setting implies a very low error tolerance, due to the high cost and small production scale.
These factors make traditional automated visual inspection techniques that rely on a ``golden sample'' approach obsolete.
Automation remains desirable, as fully manual inspection ties up qualified, expensive personnel for extended periods of time and is prone to performance degradation due to operator fatigue~\cite{huang2015automated}.

\begin{figure}[htbp]
    \centering
    \includegraphics[width=0.65\linewidth]{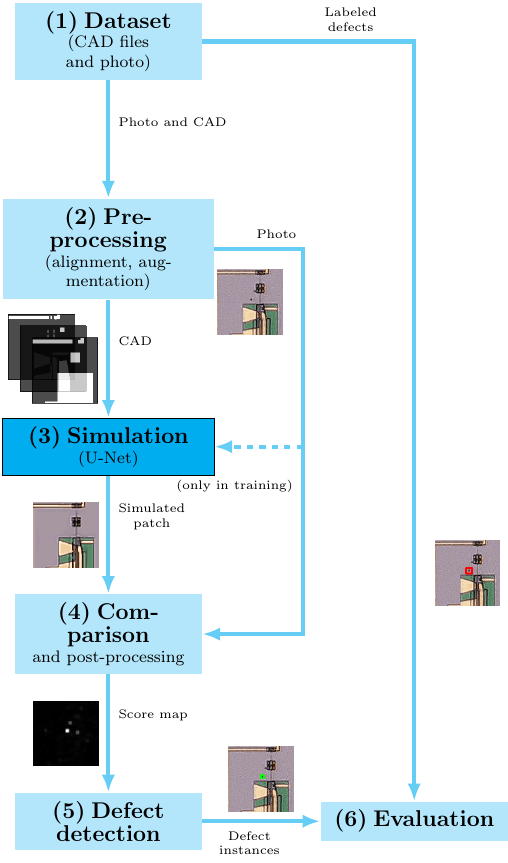}
    \caption{An overview of our proposed ``golden die'' simulation methodology.
    (1) A dataset consists of wafer image tiles with corresponding CAD segments.
    (2) After performing pre-processing steps,
    (3) a fully convolutional neural network is trained during the system's optimization phase. In production, the system then only ingests CAD segments to simulate defect-free wafer images.
    (4) Via a comparison of the simulated ``golden die`` for the CAD segment to an image of the actually produced wafer, per-pixel differences can be recorded, leading to (5) automated defect detection and scoring capabilties in experimental InP MPW.
    Prior to deployment, the defect detection performance can be evaluated against ground-truth labels (6).
    \label{fig:workflow}
    }
\end{figure}

The main challenge for an automated visual inspection approach for the MPW is the lack of ground truth labels for die inspection.
To tackle this issue, we propose a method to generate a synthetic, defect-free golden sample of a wafer from CAD manufacturing plans by training a fully convolutional neural network.
The simulation can be used for unsupervised defect detection comparing the generated image to the image of the real wafer.
An overview diagram of the proposed system is presented at Figure~\ref{fig:workflow}.

We evaluate various possible implementations for the simulation module and assess the potential for a defect detection module.
Our first approach predicts the wafer's visual representation in terms of ``material type'' as quantized color values via a segmentation model.
Our second approach uses a regression model to directly predict RGB values in pixel space. 
We discuss  and evaluate the methodological differences and practical implications of both approaches.
Both simulation outputs can then be compared algorithmically to real wafer photographs, using similarity scores from common computer vision metrics to detect and localize manufacturing defects.

The wafer layouts used in this work are small-scale prototypes, typically containing 20 to 30 unique designs, with no more than five identical wafers produced per run.
These die designs include third-party intellectual property, and labeled ground truth defect segmentation maps are generally unavailable for most wafers.
To ensure reproducibility, we publish a set of synthetic data mimicking real photonic wafers, complete with pixel-perfect labels and artificial CAD manufacturing instructions.
We evaluate our approach on both the synthetic data --- which closely matches real-world scenarios ---
and a series of real MPWs, demonstrating the practical applicability of our method.

This section first  reviews literature related to various aspects of our work.
The rest of the paper is organized as follows:
Section~\ref{data} covers the origin and processing of the data used in this project.
Section~\ref{simulation} details the simulation procedure and presents the results.
In Section~\ref{defection}  outlines a pathway for template matching-based defect detection using our simulations.
In Section~\ref{discussion} we discuss and contextualize the results.
Finally, Section~\ref{conc} summarizes our contributions, highlights the strengths and limitations of our approach, and explores potential future research.

\subsection{Related Work}

Image-based approaches for surface inspection are commonly in use in semiconductor manufacturing~\cite{bennett1995automatic}
due to their low cost, non-invasiveness \cite{xie2008review} and flexibly, as well as their potential for integration into human-operated review stations.  
In traditional industrial-scale silicon-based semiconductor manufacturing,
automatic visual inspection is easily handled by selecting a defect-free golden standard die via functional testing.
This golden sample is then used as a reference for template matching~\cite{dom1995recent, huang2015automated}.
Die-to-die inspection can also be performed to identify pairwise discrepancies between dies \cite{cho2005implementation,tsai2005eigenvalue,liu2010spectral,barone2020robust},
a method referred to as ``differential scanning''~\cite{chin1982automated}.
These approaches work when there is a large enough sample size, which is typically not the case in MPW runs.
In conventional MPW inspection, human experts manually examine the photonic integrated circuit (PIC) with a microscope~\cite{chang2011wafer,wang2006detection}.
This process is time-consuming, costly, and prone to errors due to operator fatigue, resulting in expected defect detection rates of only 60-80\%~\cite{chou1997automatic}. 

Before 2012, when deep learning-based approaches were still not dominant, state-of-the-art saliency detection methods were mostly hand-crafted \cite{zou2019object, liu2020deep}.
Today, it is common to combine learned and hand-crafted methods in the same pipeline~\cite{bhatt2021image}.
Modern approaches based on convolutional neural networks (CNNs) typically focus on supervised defect classification~\cite{chen2000neural, su2002neural, chen2007neural, chang2009application, chen2020light, kim2022novel, imoto2018cnn, cheon2019convolutional, wang2022wafer}.
This approach requires extensive ground-truth labeled data from defective and non-defective patterns.
A key challenge in object detection-based tasks, however, is the sparsity of positive samples~\cite{shrivastava2016training, lin2017focal, zhang2021weakly}, which is even more pronounced in the specific applications of defect- or anomaly detection~\cite{bhatt2021image}.
Since labeling datasets is time-consuming,
unsupervised and self- or semi-supervised approaches are desirable~\cite{chapelle2006semi},
offering significant opportunities for future research~\cite{huang2021survey}.
Examples of unsupervised surface defect detection methods include~\cite{mujeeb2018unsupervised},
which uses root mean-squared error (RMSE) in feature space to evaluate similarities between steel sheets,
and \cite{volkau2019detection},
which applies a similar method for printed circuit boards (PCBs) using a pre-trained CNN with a manually set separation threshold.
Both methods provide patch-wise scores, and divide the target image into a grid, producing a coarse defect score map.

\begin{figure*}[ht]
\captionsetup{justification=centering}
\centering
  \centering
  \includegraphics[width=0.85\linewidth]{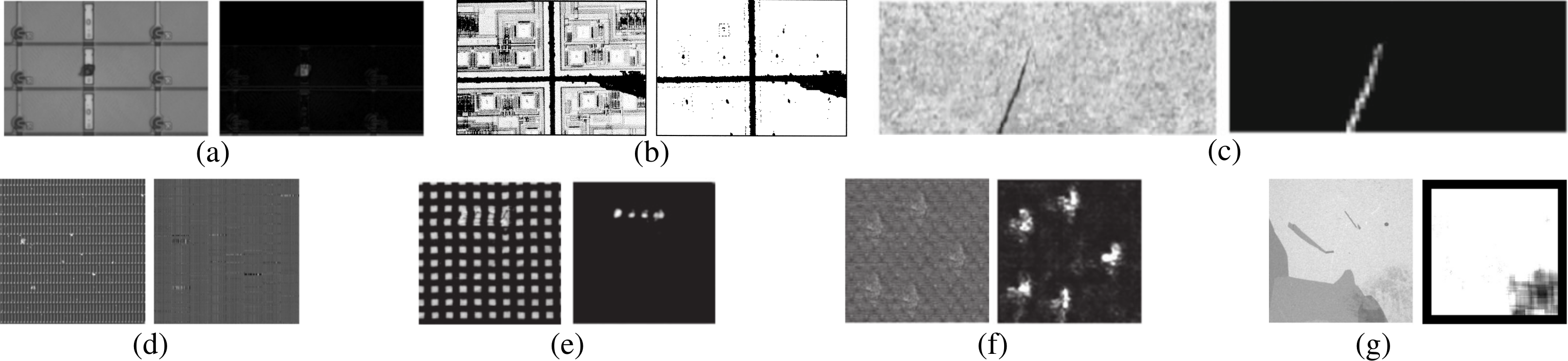}
\captionsetup{justification=justified}
\caption[Defect detection score maps in other applications]{Visualizations of the ``score maps'' generated by different defect detection methods in the literature.
On the right of each panel, the score map for the corresponding patch in the left is shown.
  (a) Defect on MVA TFT-LCD panel, from~\cite{lu2008independent}.
  (b) Die crack on electronic wafer, from~\cite{zhang1999development}.
  (c) Crack on a DC-motor commutator, from~\cite{tabernik2020segmentation}.
  (d) Particles on a TFT-LCD panel, from~\cite{lu2008independent}.
  (e) Broken ends in patterned fabric, from~\cite{hu2020unsupervised}.
  (f) Knots in patterned fabric, from~\cite{hu2020unsupervised}.
  (g) Fingerprints on solar wafer, from~\cite{li2012wavelet}.
Notice that in some domains,
such as (c) and (f), patches are very uniform and lack structure.
In contrast, the semiconductor setting typically features more intricate structures across the surface, increasing the complexity of the problem.}
\label{fig:score_maps}
\end{figure*}

Automated surface defect inspection is sought in several application domains, and is often performed by template matching.
Such methods produce a defect score map from the similarity measured between samples, such as shown in Figure~\ref{fig:score_maps}.

Our system uses a U-Net, a fully convolutional neural network, as its backbone to generate an artificial golden standard for a given wafer.
The U-Net, developed by Ronneberger et al.~\cite{ronneberger2015unet} for semantic segmentation on grayscale histological images, has an encoder-decoder architecture with skip connections that align input and output structures~\cite{isola2017image}.
The architecture has been modified for various applications across multiple fields~\cite{tao2017background, hu2019runet, liang2019concealed, kim2020color}.

Some methods applied to natural images and cityscapes generate realistic images from semantic segmentation labels~\cite{isola2017image, wang2018high, pan2019video}.
The works of \cite{wang2018high} and~\cite{isola2017image} use a cGAN to perform image-to-image translation, generating new versions of images with altered appearance by selective manipulation of classes in the labeled segmentation maps.
Results in both works are evaluated quantitatively with similarity metrics, but also qualitatively by human assessment.

To the best of our knowledge, no prior work has applied CNNs to generate a photonic wafer's golden standard in a similarly constrained setting.
While there is relevant research  in textile defect detection, motivated by similar challenges  as in semiconductors manufacturing, the key difference lies in the translation invariance of textures in textiles.
Methods in this domain  isolate recurring patterns and use a similarity metric over a golden sample~\cite{mahajan2009review, oni2018patterned}.
In particular, \cite{hu2020unsupervised} generates a golden standard from defective data using a generative adversarial network.
Unsupervised generative methods for template matching are also widely used in anomaly detection for hyperspectral aerial imagery~\cite{bati2015hyperspectral, arisoy2021unsupervised} and medical imagery~\cite{schlegl2017unsupervised}.

Evaluating results in an unsupervised setting remains a challenge, as the lack of training data (with ground truth annotations) also means that there usually is not enough evaluation data (with ground truth annotations).
Consequently, evaluating models without ground truth post-deployment requires extensive manual labor~\cite{chou1997automatic}.
This issue of evaluation without ground-truth in deployment is a known open problem~\cite{kohlberger2012evaluating,zhang2018lpips,yang2019evaluating}.

\section{Data}
\label{data}

We base our work on real wafer data but also  generate a synthetic dataset to mimic its characteristics for replicability, due to third party intellectual property constraints.
The photonic wafers in this study are manufactured through a sequence of steps, including epitaxial crystal growth, etching, deposition and lithography.
Wafer designs are stored in the GDSII format, the standard for exchanging of CAD plans for semiconductor manufacturing.
From the GDSII plans, we obtain binary bitmaps corresponding to the areas affected by each manufacturing step.
The wafers analyzed here undergo 24 manufacturing steps,
as encoded by 24 CAD bitmap layers.
To use a semantic segmentation model, the CAD plans need to be aligned with the wafer photos (described below).
This alignment process and the conversion from GDSII to matrix form is detailed in~\cite{wittmann2022photorealistic}.
We represent the binary states of the individual rasterized CAD layers with values in $\{-1,1\}^{\height \times \width \times 1}$.

The wafers are typically circular (3-inch diameter), and photographed at a resolution of 3.7$\mu$m/pixel.
The optical image $\photo$ is obtained by stitching several microscope photographs, with the final stitched image typically presenting a resolution in the order of  20k  $\times$ 20k pixels.
The stitching process is handled by the microscope's firmware\footnote{Microscope make and model: Keyence VHX-6000 with VH-ZST Zoom} and thus is not a further part of this study.
The resulting image has 3 color channels (RGB), with 8 bits per pixel, but is re-scaled and saved in our datasets as 
$\photo  \in [0,1]^{\height \times \width \times 3}$ in double precision floating point,
where $\height$ and $\width$ represent the image's height and width in pixels.
Dimensions may vary due to manual positioning on the microscope tray.

In the process described so far, both the automatic stitching from the microscope and the homographic transformation applied to align the CAD layers over the photos are sources for potential misalignment, which must be addressed downstream.

\begin{figure}[ht]
    \captionsetup{justification=centering}
    \centering
    \begin{subfigure}[t]{0.4\linewidth}
        \centering
        \includegraphics[width=0.95\linewidth]{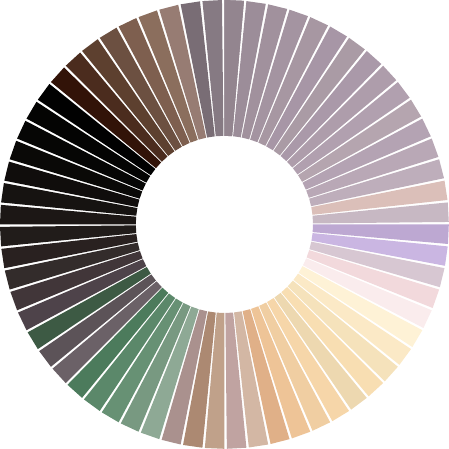}
        \caption{Real wafer.}
        \label{fig:palette_real}
    \end{subfigure}%
    \begin{subfigure}[t]{0.4\linewidth}
        \centering
        \includegraphics[width=0.95\linewidth]{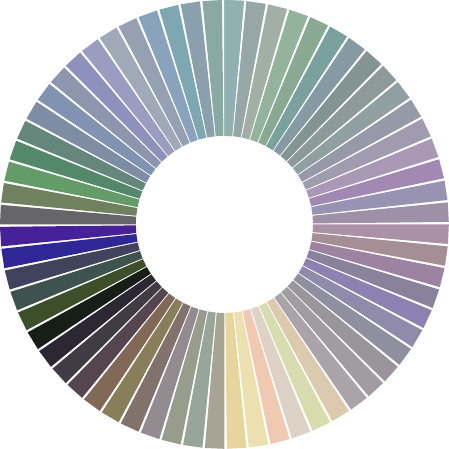}
        \caption{Synthetic wafers.}
        \label{fig:palette_mock}
    \end{subfigure}
    \captionsetup{justification=justified}
    \caption{Quantized palette samples from a real wafer (\ref{fig:palette_real}) and the synthetic wafer dataset (\ref{fig:palette_mock}).
    The palette was ordered by solving a traveling salesman problem minimizing the euclidean distance of a path passing through the 64 RGB coordinates.
    The approximate cyclic ordering of the colors allows for evaluation of \emph{off-by-one} misclassifications.
    Image best seen in color.}
    \label{fig:palette}
\end{figure}

Next to simulations targeting the complete range of the full color RGB color space, we also generate and evaluate simulations in a reduced, quantized color-space.
To this end, each wafer image is represented as a single-channel 64-color quantized image $\photo_q \in \mathbb{Z} \cap [0..63]^{\height \times \width}$,
where the pixel value corresponds to the index of a color in the wafer's color palette inferred from quantization.
The quantization process, described in \cite{dolgenercantu2023image}, reduces the $256^3$ RGB colors to 64 centroids obtained through $k$-means clustering of a sample of the pixels from the wafer.
See Figure~\ref{fig:palette} for example color palettes obtained from a real wafer photograph and synthetically generated wafer images.

\begin{figure}[ht]
    \centering
    \begin{subfigure}[t]{0.25\linewidth}
      \centering
      \includegraphics[width=0.95\linewidth]{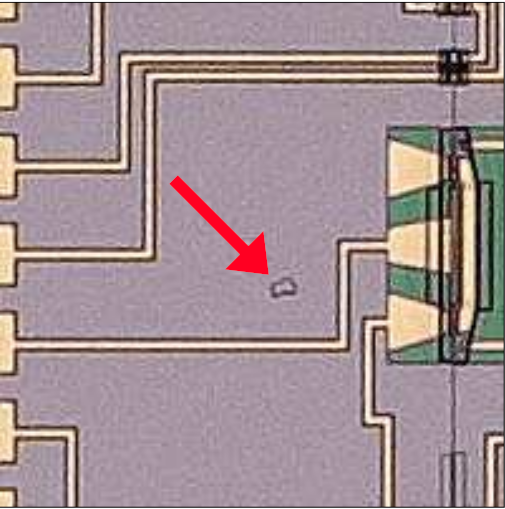}
      \caption{Wafer photo.}
      \label{fig:data1}
    \end{subfigure}%
    \begin{subfigure}[t]{0.25\linewidth}
      \centering
      \includegraphics[width=0.95\linewidth]{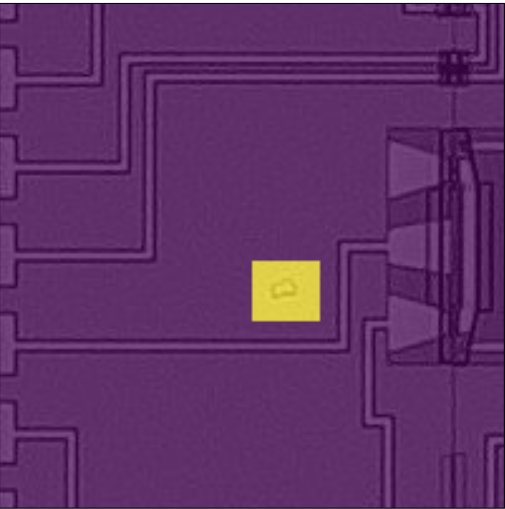}
      \caption{Label mask.}
      \label{fig:data2}
    \end{subfigure}%
    \begin{subfigure}[t]{0.25\linewidth}
      \centering
      \includegraphics[width=0.95\linewidth]{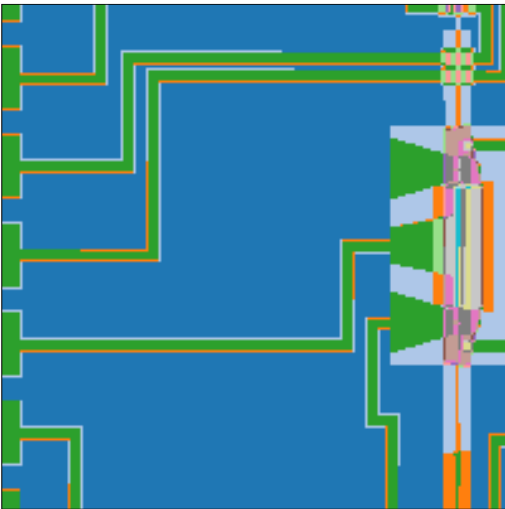}
      \caption{CAD layers.}
      \label{fig:data3}
    \end{subfigure}
    \caption[Our dataset]{Visual representation of various data types associated with a wafer.
    Defect labels (see (b)), stored as a binary map in $\{-1,1\}$, are overlaid on the photograph (a) for clarity.
    Defects (see red arrow in (a)) are typically smaller than the corresponding labels (yellow in (b)), which also encompass non-defective regions. 
    The final image colorizes stacked CAD layers, each assigned a unique color. These layers inform about InP wafer manufacturing steps, are stored as binary images, and may overlap due to multiple processing steps affecting the same wafer area.
    \label{fig:data}
    }
\end{figure}

The CAD building plans and corresponding wafer images serve as inputs and training targets for the simulator network, respectively.
Finally, the dataset contains a small amount of manually labeled defects.
Their utility for training is limited, though, as only one of the designs in one of the wafers has been completely labeled.
Due to the high cost and time consumption of the labeling task, the available labels are very sparse, and provide relatively coarse masks --- i.e. bounding boxes covering an area larger than the actual defect's footprint in the image.
The labels are gathered using a custom designed software annotation tool, with which a user draws the contours for a certain defect and assigns a defect class to the marked area.
Defect class information can be of further use in defect classification tasks and is present where possible in the dataset.
For this work, labels are processed as binary masks, as we focus on defect localization only.
The detailed description of the data used is presented in previous work \cite{dolgenercantu2023image}, and a sample patch in our datasets contains the layers shown in Figure~\ref{fig:data}.

The results shown in this work originate mostly from a set of generated synthetic datasets, for two reasons:
the data pertaining to real wafers contains third party intellectual property, which the authors have no permission to reproduce.
The artificial data further allows us to parametrize defect density for a few classes of defects.
Our artificial data is analog to the real data, with the files following the same structure, with the difference that it has only 5 layers in the GDS data.
The models trained with one set of data are thus not directly usable for the other type of data.
The artificial dataset will be publicly available for reproducibility upon acceptance of this work.
In the following experiments, we present results for both variants of data.

\begin{figure}[!h]
    \centering
    \includegraphics[width=.6\linewidth]{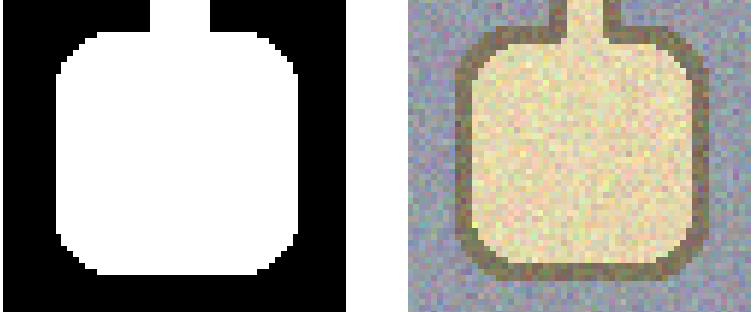}
    \caption[Toy data - contextual dependence]{
    A sample pair of a single CAD layer and corresponding image in our synthetic dataset.
    The remaining 4 CAD layers are empty.
    The borders of the component have a distinct color in the region between substrates.
    The local context is not explicit in the binary CAD mask, making such samples a suitable training pair.}
    \label{fig:toydata_context}
\end{figure}

As reference synthetic data for training we generated five datasets of square images of 10k $\times$ 10k pixels in size.
In CAD layers corresponding to textual information expressed on the wafers, approximately 20\% of the characters contain defects.
The remaining classes of defects -- point defects, residual resists and nitride liftoff -- are present at an approximate rate of 8 defects/MPx, for each type.
Such defect densities are perceptually similar to what is typically observed in the real data.
All defects are accompanied with pixel perfect ground truth localization.

\begin{figure}[ht]
    \centering
  \includegraphics[width=.88\linewidth]{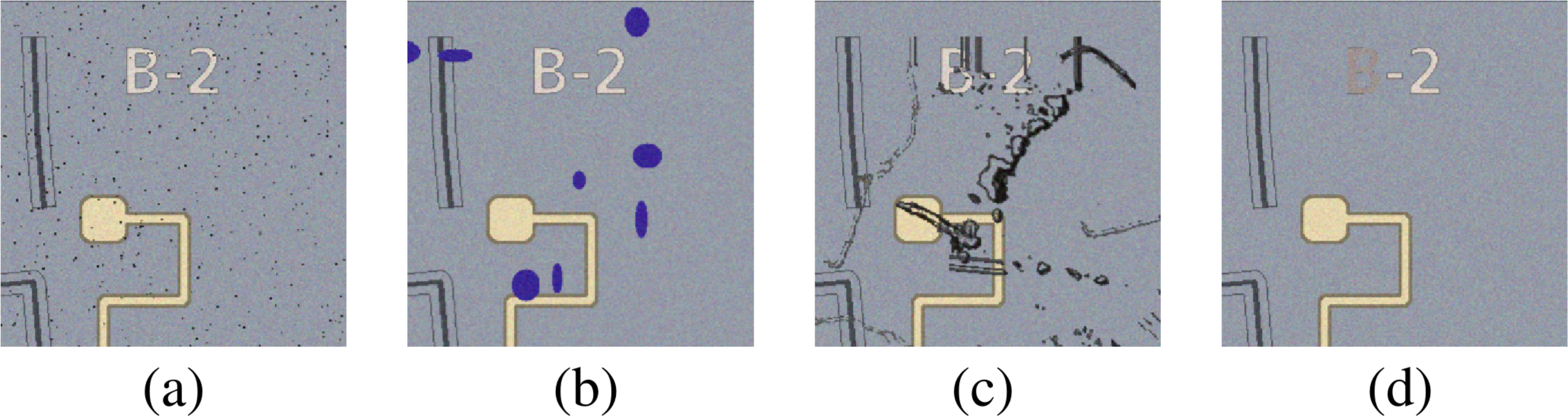}
    \caption[Toy data - defect classes]{The 4 defect classes implemented in the toy data:
    (a) dust particles, at 10,000 defects/MPx;
    (b) nitride liftoff, at 100 defects/MPx;
    (c) residual resists or strange particles, at 100 defects/MPx;
    (d) burned platinum affecting 50\% of the letters.
    The samples shown here were generated with a disproportionally high defect density when compared to real data, for increased visibility.
    }
    \label{fig:toydatadefectclasses}
\end{figure}

\section{Simulation of an Artificial Golden Standard}
\label{simulation}

Our primary objective is to determine whether it is possible to simulate a perceptually photorealistic wafer without defects, even when the training data contains imperfections.
The location of the manufacturing defects is a-priori unknown.
Due to the sparsity of these defects in the wafer photographs and the clear correspondence from the structures in the CAD layers to the desired features in the photograph, it is \emph{expected} that the simulations will not systematically replicate those defects and anomalies.

We train a model $U$, a U-Net which approximates 
$$ \hat{\mathbf{p}}(\mathcal{R}) = U(\mathbf{g}(\mathcal{R}))$$ from
the appearance of a sample patch $\mathbf{p}(\mathcal{R}) \in \photo$, the photographed wafer, based on the $\mathbf{g}(\mathcal{R})$ CAD layouts in the corresponding rectangular region
$$\mathcal{R} = \{(x, y) \mid x_{i} \leq x \leq x_{i+w}, \, y_j \leq y \leq y_{j+h}\}.$$

We adapt the original U-Net structure described by \cite{ronneberger2015unet} as needed.
The network's input is the CAD layers $\mathbf{g} \in \mathbb{R}^{ k_\text{in} \times h \times w}$, for $k_\text{in}$ CAD layers, where $h$ and $w$ are height and width of the simulated patch, respectively.
For synthetic data, $k_{in}=5$, and for real wafers it varies according to the specific number of steps in the foundry's process.
The convolutions are zero-padded, eliminating cropping and providing outputs with the same resolution as the layer's input.
We use batch normalization \cite{ioffe2015batch} after each convolutional layer, which is also adopted in a subsequent presentation of the original U-Net \cite{cciccek20163d}.
The upsampling is implemented through bilinear interpolation, which does not contain trainable parameters.
The output of the last convolution layer is $\hat{\mathbf{x}} \in \mathbb{R}^{k_\text{out} \times h \times w}$, where $k_\text{out}$ denotes the number of output channels.

The training objective is to minimize the dissimilarity $ S(\hat{\mathbf{p}}, \mathbf{p})$
between the predicted and the actual wafer photographs.
We train models for \textit{regression} of the exact pixel values and per pixel \textit{classification} models in quantized color space.
The loss function $S$ is chosen according to the learning task.

For the regression in RGB space, we have $k_\text{out} = 3$, for the red, green and blue (RGB) channels.
The logits $\hat{\mathbf{x}}$ are scaled by the elementwise scaled hyperbolic tangent transformation
\begin{equation}
 \hat{\mathbf{p}}_\text{RGB} = \frac{\text{tanh}(\hat{\mathbf{x}})}{2}+0.5, 
 \end{equation}
which constraints the output $\hat{\mathbf{p}}_\text{RGB} \in [0,1]^{3 \times h \times w}$.

For the classification models, which use $k_\text{out} = 64$ and a discretized color palette $\mathbf{q} \in [0,1]^{3 \times k_\text{out}}$.
From the vector $\hat{\mathbf{x}}_{hw} \in \mathbb{R}^{k_{out}}$ of class scores for a certain pixel we map the RGB representation of a simulated pixel $\hat{p}_{bhw} \in \mathbf{\hat{p}}_\text{RGB}$ with 
 \begin{equation}
 \hat{p}_{bhw} = \mathbf{q}_k \text{, where } k = \argmax_{[1,k_{out}]} \hat{\mathbf{x}}_{bhw}.
 \end{equation}

We train unique models for each different wafer, as each wafer run might present very distinct visual characteristics from one another, as well as different orderings in the color palettes.
Therefore, each model can be used solely with data from the same wafer run that originated the training data.

\subsection{Training details}

In the regression case, we train a model on each dataset to minimize the learned perceptual image patch similarity (LPIPS) \cite{zhang2018lpips} and another series of models minimizing the mean squared error (MSE) as target function.
For the classification tasks, the objectives used are the categorical cross entropy, which is widely used for semantic segmentation and classification problems, and the focal loss, which is an expansion of cross entropy with reportedly good results on unbalanced datasets \cite{lin2017focal}.
For each of the 4 objective functions used, we train a model on each of the 9 real wafers available and for each of 5 sets of synthetic data.
Additionally, we train a decision tree for benchmarking on each of the 14 datasets.
We present numerical evaluation results for all models, yet our presentation of visual samples is limited due to third-party content on the wafers and copyright considerations.

For training, we use patches 64 $\times$ 64 pixels in size, and batches have at most 128 patches.
A considerable amount of patches in the wafer depict mostly uniform regions with no functional components --- and even component-dense patches show a considerable portion of background.
To balance the exposure of the model during training to the different kinds of visual features, we disconsider samples which do not reach a certain variance threshold in the quantized target sample.
It is interesting to note that the eventual smaller batch sizes potentially accelerates the training process.
However, a comprehensive investigation into these effects falls beyond the scope of the present work and will not be analyzed.
In our experimental setup, we trained models across a variety of thresholds, specifically ranging from 0 to 50.
See \ref{ap:training} for details on which thresholds were used in each model.
An initial learning rate of $5\times10^{-3}$ was adopted for all trained models.
A decay implemented as a multiplicative factor of 0.6 is applied after epochs 1, 3, 5 and 8.
Following the 8th epoch, learning rate is kept at a constant value of $6.48\times10^{-4}$.
We use stochastic gradient descent as weight update rule.

All models were trained on a 70\% partition of the corresponding dataset.  
No input transformations or augmentations are applied to the training samples.
The validation of models trained on real wafers was performed on the unseen remaining 30\% subset of the respective wafer.
Each model trained on synthetic data is cross-validated on the the remaining datasets not employed in its own training.

\subsection{Evaluation methodology}

Our simulations are assessed based on their similarity to the original image and their ability to reject defects.  
The quality of simulations is quantitatively evaluated, with reported similarity metrics comparing simulated patches to original wafer images.  
Regarding defect rejection, it is acknowledged that simulation targets possess inherent unlabeled imperfections, and purely quantitative metrics may not adequately capture perceptual similarity or the nuances of failure modes.  
Consequently, a qualitative analysis of the results is also presented.

We present results for the models generated after the 10-th training epoch and for the best performing intermediate checkpoints.
We select a set of \textit{best models} according to the MSE similarity to the ground truth and a set of checkpoints based on their LPIPS performance.
This is approach is beneficial given that both metrics might not necessarily be correlated.

To provide a reference performance benchmark we also train decision trees in the quantized space on a random sample of $5\times10^6$ pixels for each of the available datasets.
The training samples are randomly ordered pixels paired with the corresponding CAD layer values, $\{p_{hw}, \mathbf{g}_{hw}\}$.
Due to the pixelwise approach, the trees do not learn contextual information -- in contrast to the patchwise input adopted for the neural networks, which integrates neighboring pixel information into the final prediction via convolutional filters.

We evaluated the simulation outcomes by comparing the performance of trained models across different wafers and reported patchwise averages and standard deviations for a few similarity metrics on the unseen validation data.
Inference is performed with batches of 20 square samples of $256 \times 256$ pixels.
With the addition of the decision tree baseline models, we provide evaluation results for 167 models.

A global metric for an entire wafer can be derived by gathering the results from all simulated patches or even approximated by measures from a subset of the sample space.
We compute for all models metrics in RGB space, either directly from the RGB images or from the reconstructed quantized RGB images for classification models.
In addition, the classification models were also evaluated with semantic segmentation metrics (cross entropy, $k$-off accuracy and focal loss), comparing the categorical index of the quantized color in the target photo with the model's prediction confidence across classes.
For pixelwise metrics, the aggregated average for each patch is reported.  
Remaining metrics -- SSIM, LPIPS, HaarPSI and PSNR -- yield patchwise results, and therefore reported results consist of mean values across all patchwise results and standard deviation across patches.
We present the most significant metrics in the section below, and the complete tables of results can be seen on \ref{ap:results}.

\subsection{Results}

\begin{figure}[ht]
\centering
\includegraphics[width=0.75\linewidth]{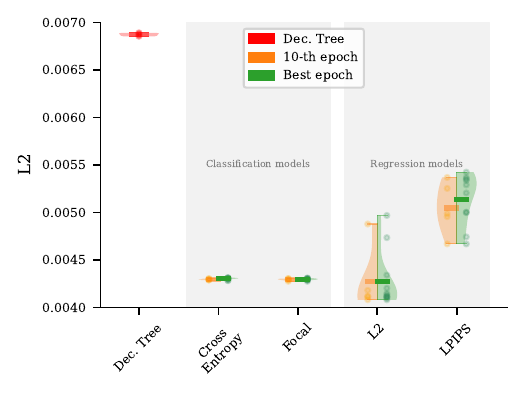}
\caption{L2-norm validation loss for the models trained on synthetic data, grouped by which objective function was used for training.
Lower is better.
The training of classification models is more stable than of regression models.
Even then, the quantitative performance of most models surpasses the tree's performance.}
\label{fig:l2_mock_plot}
\end{figure}

\begin{figure}[ht]
\centering
\includegraphics[width=0.75\linewidth]{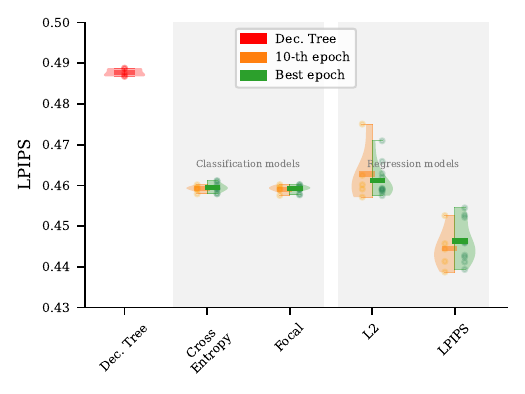}
\caption{LPIPS validation loss for the models trained on synthetic data, grouped by which objective function was used for training.
Lower is better.
LPIPS provides scores with a higher variance and ranks models differently than L2, but confirms the same performance trends.
The \textit{best epochs} were also selected due to their L2 measure, and we can see that the best L2-scored models are not necessarily also optimized for LPIPS.
The quantitative performance of most models surpasses the tree's performance.}
\label{fig:lpips_mock_plot}
\end{figure}

The violin plots in Figures~\ref{fig:l2_mock_plot} and~\ref{fig:lpips_mock_plot} illustrate the performance of models trained on synthetic data, grouped by objective function used in training.
Points in both plots show the same models, however Figure~\ref{fig:l2_mock_plot} ranks these models according to the L2 distance between predictions and ground truth target, while Figure~\ref{fig:lpips_mock_plot} evaluates simulations based on the LPIPS metric.
It is evident that all models overperform the baseline offered by the decision trees,
and that results demonstrate a notable consistency among the classification models.
While regression models also exhibit considerable capability, it is necessary to care for selection of a high-performing intermediate checkpoint.
It also is noteworthy that the best performing models in terms of the L2 distance do not necessarily rank highly in the LPIPS sense (and vice versa), indicating the importance of careful selection of training targets and evaluation metrics.
All observed models yield realistic simulations.

The results shown on Figures~\ref{fig:l2_mock_plot} and~\ref{fig:lpips_mock_plot}, however, are inconclusive on indicating which target function is generally more appropriate, as models tend to perform better when evaluated on the same function used as a training objective.

\begin{figure}[ht]
\captionsetup{justification=centering}
\centering
\begin{subfigure}[t]{0.25\linewidth}
  \centering
  \includegraphics[width=0.92\linewidth]{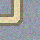}
  \caption{Target sample.}
  \label{fig:mock_patches1}
\end{subfigure}%
\begin{subfigure}[t]{0.25\linewidth}
  \centering
  \includegraphics[width=0.92\linewidth]{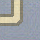}
  \caption{U-Net simulation.}
  \label{fig:mock_patches2}
\end{subfigure}%
\begin{subfigure}[t]{0.25\linewidth}
  \centering
  \includegraphics[width=0.92\linewidth]{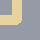}
  \caption{Decision tree.}
  \label{fig:mock_patches3}
\end{subfigure}%
\captionsetup{justification=justified}
\caption{Samples of a simulation on synthetic data.
The simulator is a regression model trained with LPIPS as objective function.
For the patch pictured, the U-Net scored L2 0.0032 and LPIPS 0.02, while scoring L2 0.005 and LPIPS 0.446 for the whole wafer.
The tree scored L2 0.0147 and LPIPS 0.08 on this patch, and L2 0.007 and LPIPS 0.486 on the whole wafer.
The U-Net was trained on LPIPS loss.}
\label{fig:mock_patches}
\end{figure}

\begin{figure}[ht]
\captionsetup{justification=centering}
\centering
\begin{subfigure}[t]{0.25\linewidth}
  \centering
  \includegraphics[width=0.92\linewidth]{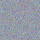}
  \caption{Target sample.}
  \label{fig:mock_patches_bg_1}
\end{subfigure}%
\begin{subfigure}[t]{0.25\linewidth}
  \centering
  \includegraphics[width=0.92\linewidth]{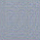}
  \caption{U-Net simulation.}
  \label{fig:mock_patches_bg_2}
\end{subfigure}%
\begin{subfigure}[t]{0.25\linewidth}
  \centering
  \includegraphics[width=0.92\linewidth]{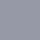}
  \caption{Decision tree.}
  \label{fig:mock_patches_bg_3}
\end{subfigure}%
\captionsetup{justification=justified}
\caption{Samples of the prediction on mock data, purely background.
On this patch, the U-Net scores L2 0.0031 and LPIPS 0.13.
For the whole wafer, it scores L2 0.005 and LPIPS 0.446.
The decision tree shows L2 0.007 and LPIPS 0.486 for this patch,
and L2 0.0023 and LPIPS 0.42 for the whole wafer.
The U-Net was trained on LPIPS loss.}
\label{fig:mock_patches_bg}
\end{figure}

We present in Figure~\ref{fig:mock_patches} and Figure~\ref{fig:mock_patches_bg} predictions generated by a model trained on the regression task, its decision tree-based counterpart and the target patch they are simulating.
These samples show that the scores presented by the models are in a range that is well suited for the problem at hand.
Figure~\ref{fig:mock_patches3} depicts how the decision tree fails to capture the contextual information present on the transition from the traces to the substrate background, as expected.
The CAD layers for this component show only a binary value where the trace runs, and the manufacturing process introduces the darker areas on the edges of the trace, which are reproduced adequately on Figure~\ref{fig:mock_patches2}.

The patch on Figure~\ref{fig:mock_patches_bg} covers a region that shows only the wafers substrate background.
Both the prediction from the tree and the one from the neural network are perceptually similar, and the decision tree-based simulation scores better in L2 similarity.
The neural network prediction from Figure~\ref{fig:mock_patches_bg_2} shows the network's capability of learning noise-based features contained in the photo, such as the random noise, and scores better with LPIPS than the decision tree-generated sample.

\begin{figure}[ht]
\centering
\includegraphics[width=0.75\linewidth]{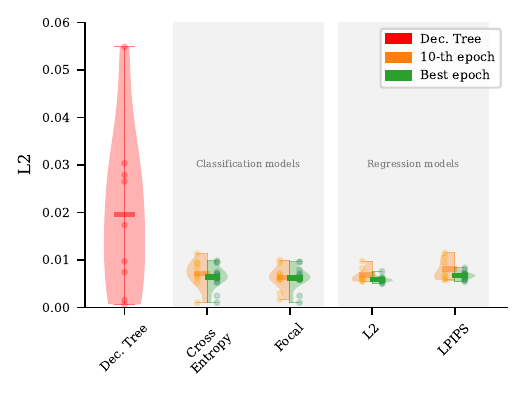}
\caption{L2-norm validation loss for the models trained on real data.
They are grouped by their type of data and training objective used.
Lower is better.
For a better visualization, there are three datapoints omitted from the plot, in the set of 10-th epoch regression models.
The mean performance obtained is still correctly plotted.
It is harder to achieve consistent models training on real data than on synthetic data, both for the U-Net and for decision trees.
Cross-entropy loss on a classification setting showed better overall consistency in the experiments carried on this work.}
\label{fig:l2_real_plot}
\end{figure}

\begin{figure}[ht]
\centering
\includegraphics[width=0.75\linewidth]{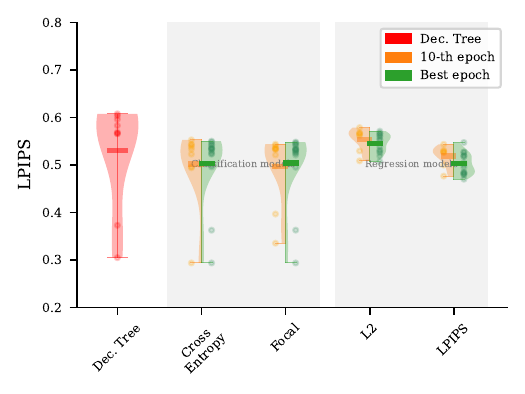}
\caption{LPIPS validation loss for all of the models trained on real MPW wafers.
They are grouped by their type of data and training objective used.
Lower is better.
It is harder to achieve consistent models training on real data, both for the U-Net and for decision trees.
Cross-entropy loss on a classification setting showed better overall consistency in the experiments carried on this work.}
\label{fig:lpips_real_plot}
\end{figure}

Figures~\ref{fig:l2_real_plot} and~\ref{fig:lpips_real_plot} present the performance of models trained on real MPW wafers.
The real wafer datasets exhibit substantially greater visual variability compared to the synthetic datasets, characterized by diverse color palettes, design layouts, and structural features.
As illustrated in Figure~\ref{fig:l2_real_plot}, the benchmark decision tree models yield more heterogeneous results, with L2-norm values approximately an order of magnitude higher than those observed for models trained on synthetic data.
Consistent with observations from the synthetic data analysis, results from regression models tend to exhibit greater variability across runs compared to classification models.
Nonetheless, after filtering for high-performing intermediate checkpoints, regression models are capable of achieving competitive performance.

\begin{figure}[ht]
\captionsetup{justification=centering}
\centering
\begin{subfigure}[t]{0.25\linewidth}
  \centering
  \includegraphics[width=0.92\linewidth]{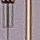}
  \caption{Target sample.}
  \label{fig:real_patches1}
\end{subfigure}%
\begin{subfigure}[t]{0.25\linewidth}
  \centering
  \includegraphics[width=0.92\linewidth]{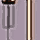}
  \caption{U-Net simulation.}
  \label{fig:real_patches2}
\end{subfigure}%
\begin{subfigure}[t]{0.25\linewidth}
  \centering
  \includegraphics[width=0.92\linewidth]{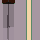}
  \caption{Decision tree.}
  \label{fig:real_patches3}
\end{subfigure}%
\captionsetup{justification=justified}
\caption{Samples of the prediction on real data.
The U-Net based simulation scored L2 of 0.009 and LPIPS of 0.52 in the whole wafer; and L2 of 0.0132, LPIPS of 0.04 for the depicted patch.
The decision tree for this wafer scored L2 0.055 and LPIPS 0.61, while scoring L2 0.0147 and LPIPS 0.06 for this specific patch.
The U-Net was trained on cross-entropy loss.}
\label{fig:real_patches_a}
\end{figure}

\begin{figure}[ht]
\captionsetup{justification=centering}
\centering
\begin{subfigure}[t]{0.25\linewidth}
  \centering
  \includegraphics[width=0.92\linewidth]{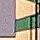}
  \caption{Target sample.}
  \label{fig:real_patches1b}
\end{subfigure}%
\begin{subfigure}[t]{0.25\linewidth}
  \centering
  \includegraphics[width=0.92\linewidth]{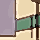}
  \caption{U-Net simulation.}
  \label{fig:real_patches2b}
\end{subfigure}%
\begin{subfigure}[t]{0.25\linewidth}
  \centering
  \includegraphics[width=0.92\linewidth]{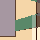}
  \caption{Decision tree. }
  \label{fig:real_patches3b}
\end{subfigure}%
\captionsetup{justification=justified}
\caption{Samples of the prediction on real data.
Whole wafer: L2 0.009, LPIPS 0.52 patch: L2: 0.0311, LPIPS: 0.03 for neural net.
Whole wafer: L2 0.055, LPIPS 0.61 patch L2: 0.0272, LPIPS: 0.04  for tree.
Notice that for this patch LPIPS is better for the neural net-based simulation, while the L2 score is better for the decision tree model.
The U-Net was trained on cross-entropy loss.}
\label{fig:real_patches_b}
\end{figure}

When evaluated on LPIPS performance, which is constrained in the interval $[0,1]$, models trained on real data again show more variability than the ones trained on synthetic data, with slightly worse average performance, as shown by the plot in Figure~\ref{fig:lpips_real_plot}.
The perceptual results depicted in Figures~\ref{fig:real_patches_a} and~\ref{fig:real_patches_b} confirm qualitatively that the range of scores presented by the simulations relates to realistic outputs.
The dichotomy between perceptual quality and measured similarity shows also in Figure~\ref{fig:real_patches_b}, where the tree-based simulation scores better than its neural net-based counterpart, despite the latter exhibiting greater level of detail. 
Furthermore, no significant differences were detected between cross-entropy loss and focal loss in terms of training duration or the performance achieved.

\begin{figure}[ht]
\captionsetup{justification=centering}
\centering
\begin{subfigure}[t]{0.16\textwidth}
  \centering
  \captionsetup{width=.8\linewidth}
  \includegraphics[width=.9\linewidth]{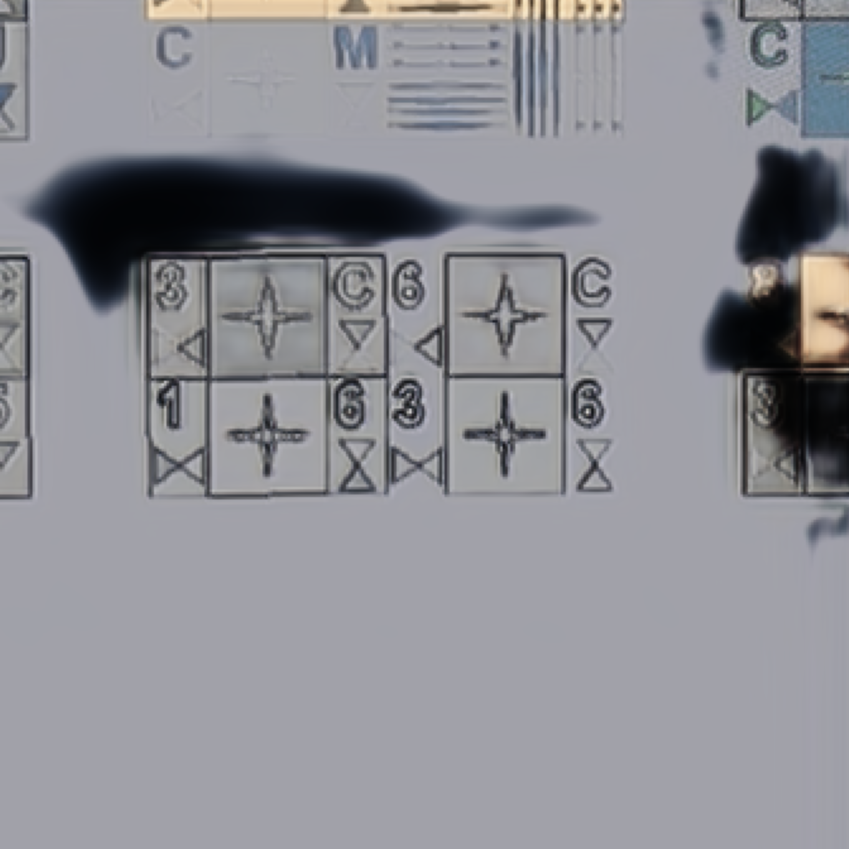}
  \caption[width=0.1\linewidth]{LPIPS.}
  \label{fig:hallu_sample1}
\end{subfigure}%
\begin{subfigure}[t]{0.16\textwidth}
  \centering
  \captionsetup{width=.8\linewidth}
  \includegraphics[width=.9\linewidth]{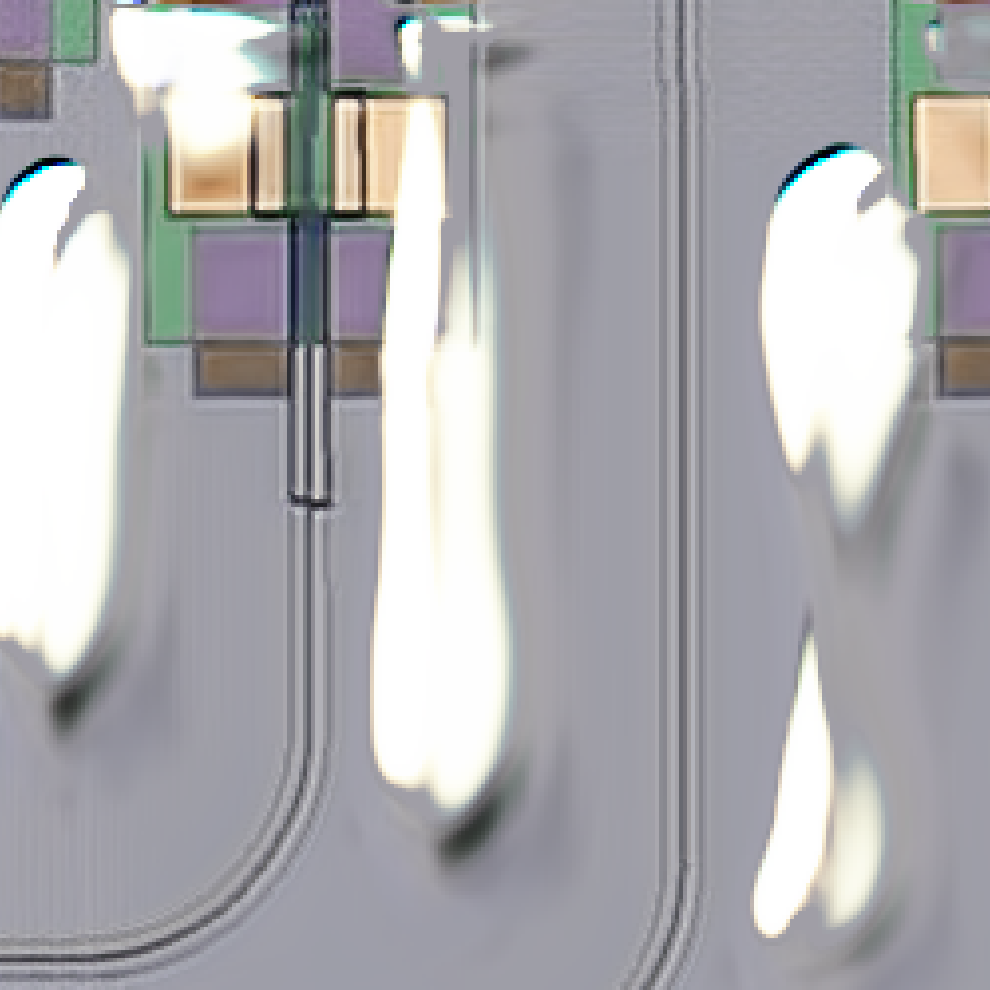}
  \caption{LPIPS.}
  \label{fig:hallu_sample2}
\end{subfigure}%
\begin{subfigure}[t]{0.16\textwidth}
  \centering
  \captionsetup{width=.\linewidth}
  \includegraphics[width=.9\linewidth]{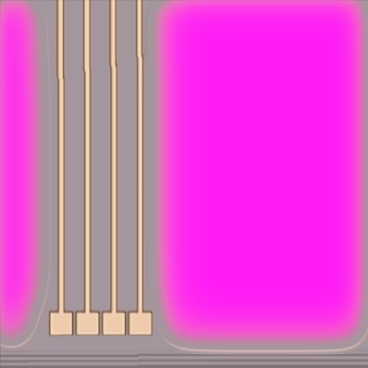}
  \caption{L2.}
  \label{fig:hallu_sample3}
\end{subfigure}%
\begin{subfigure}[t]{0.16\textwidth}
  \centering
  \captionsetup{width=.9\linewidth}
  \includegraphics[width=.9\linewidth]{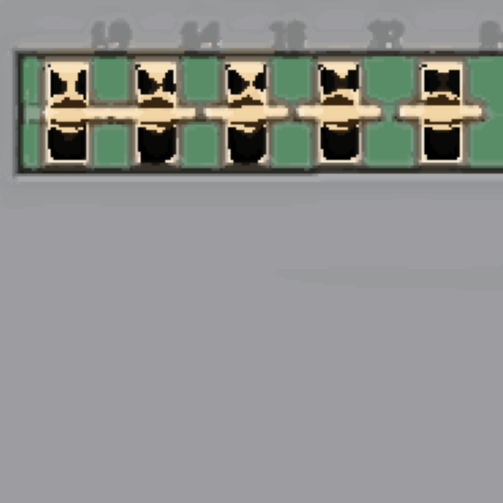}
  \caption{Cross entropy.}
  \label{fig:hallu_sample4}
\end{subfigure}%
\begin{subfigure}[t]{0.16\textwidth}
  \centering
  \captionsetup{width=.7\linewidth}
  \includegraphics[width=.9\linewidth]{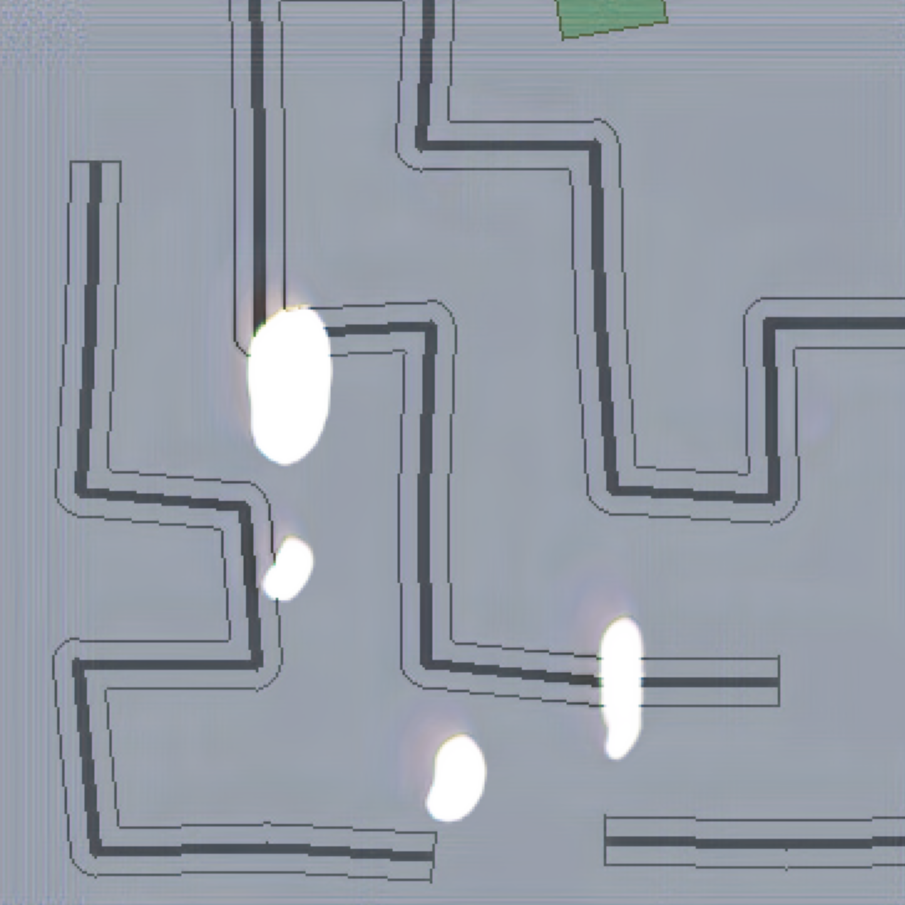}
  \caption{LPIPS.}
  \label{fig:hallu_sample5}
\end{subfigure}%
\begin{subfigure}[t]{0.16\textwidth}
  \centering
  \captionsetup{width=.7\linewidth}
  \includegraphics[width=.9\linewidth]{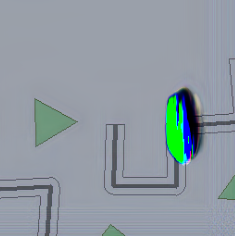}
  \caption{LPIPS.}
  \label{fig:hallu_sample6}
\end{subfigure}
\captionsetup{justification=justified}

\caption{Simulations from models trained on the indicated target function.
Some models learn to hallucinate artifacts.
Those artifacts are not similar to any structures contained in the wafer, and are usually covering background areas or uniformly colored areas. Most of the hallucinations present themselves as a saturation in one or more of the channels.
The caption in each of the samples indicates the target function used in training.}
\label{fig:bad_patches}
\end{figure}

Even though the proposed method is demonstrated to be robust in obtaining a faithful simulator with a relatively short training run, it remains susceptible to failure modes, which we report here.
In certain cases, the learned models produce hallucinations and artifacts within the simulation.
Examples of this effect are illustrated in Figures~\ref{fig:bad_patches}.
The averaged patch-wise validation loss for the depicted models scored satisfactorily across the whole wafer, indicating that simulation success is not always clearly reflected by the validation loss and that averaged patch-wise performance metrics may be insufficient for reliably detecting such anomalies at a wafer level.

\begin{figure}[ht]
\captionsetup{justification=centering}
\centering
\begin{subfigure}[t]{0.25\linewidth}
  \centering
  \includegraphics[width=0.98\linewidth]{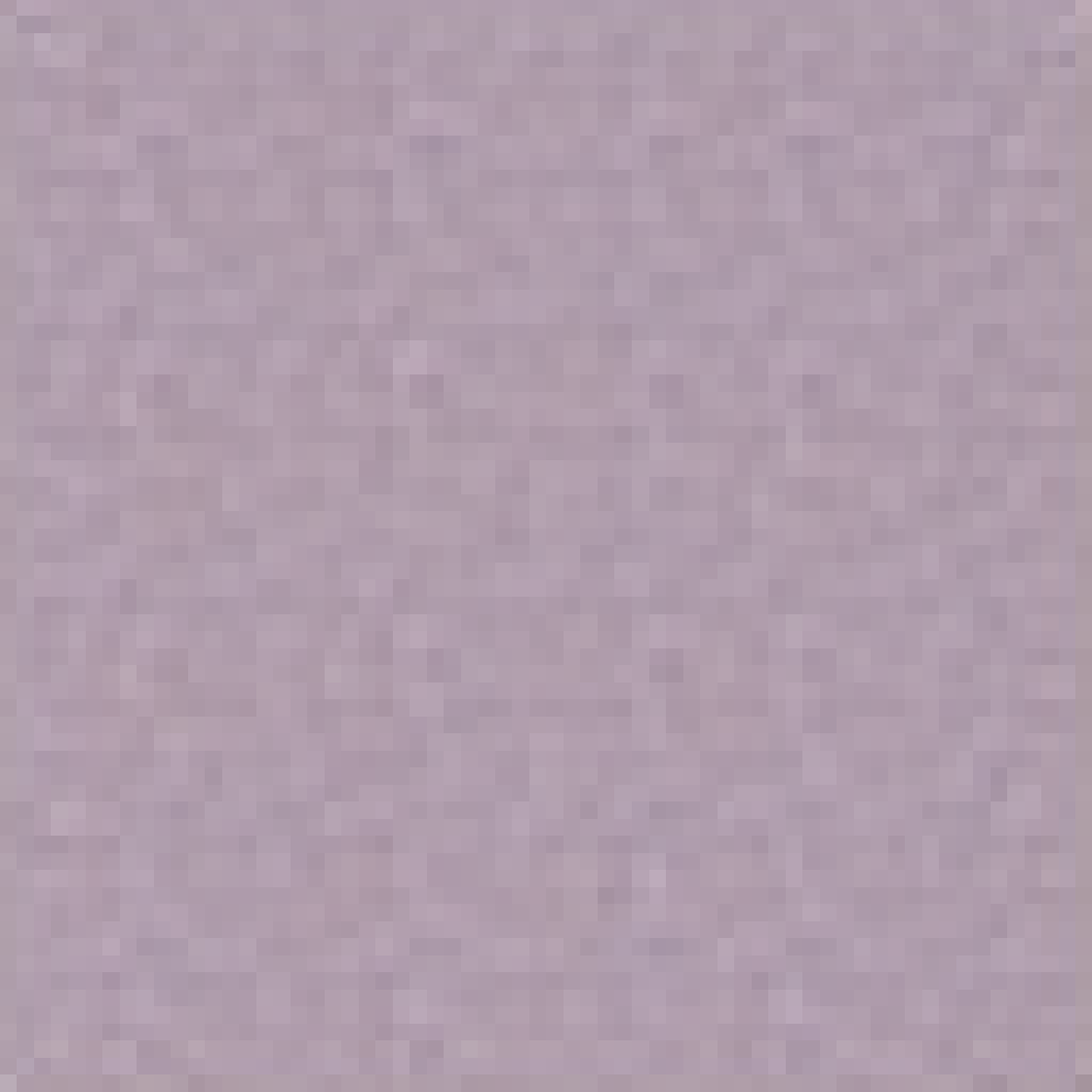}
  \caption{64 $\times$ 64}
  \label{fig:bg_ripple_64}
\end{subfigure}%
\begin{subfigure}[t]{0.25\linewidth}
  \centering
  \includegraphics[width=0.98\linewidth]{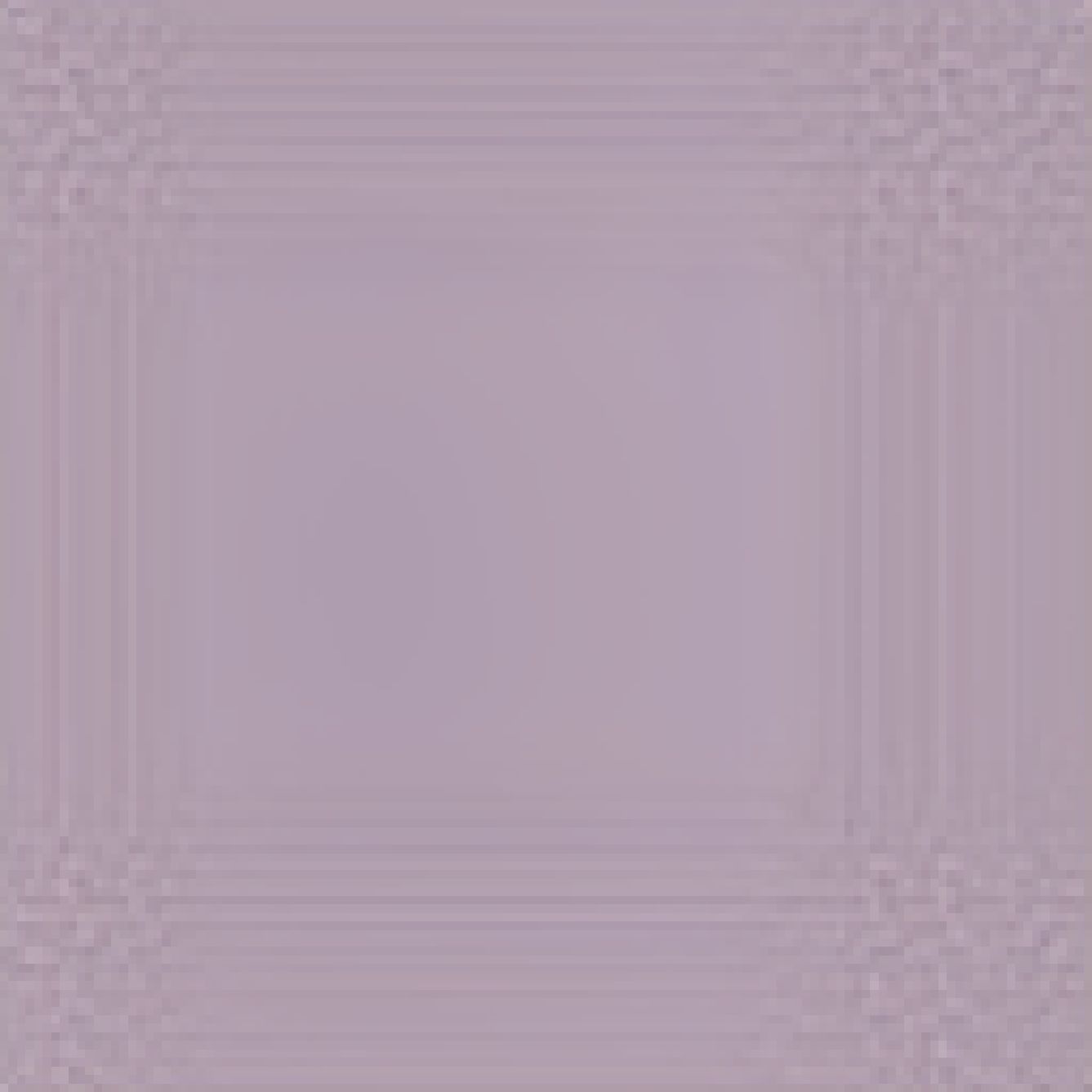}
  \caption{128 $\times$ 128}
  \label{fig:bg_ripple_128}
\end{subfigure}%
\captionsetup{justification=justified}
\caption{Appearance of the ripples observed.
We show a background patch for ease of visualization.
The model which generated the depicted samples was trained on 64$\times$64 pixel samples,
using LPIPS as a loss function.}
\label{fig:bg_ripple}
\end{figure}

A noteworthy effect is observed in models trained specifically using LPIPS as a loss function.  
Namely, the simulations generated by these models exhibit artifacts resembling \emph{waves} or \emph{ripples} along the periphery of the simulated patches when the input sample size exceeds that used during training.
Such effect can be observed in Figures~\ref{fig:mock_patches_bg_2},~\ref{fig:mock_patches2} and~\ref{fig:hallu_sample6}.
Figure~\ref{fig:bg_ripple} highlights this phenomenon in two simulated examples: one generated from a 64$\times$64-pixel CAD input (Figure~\ref{fig:bg_ripple_64}) and another from a 128$\times$128-pixel input (Figure~\ref{fig:bg_ripple_128}).

We hypothesize that the emergence of this undulating structure is a result of the three successive downsampling operations in the U-Net's convolutional architecture.
In Figure~\ref{fig:bg_ripple_128} the superposition of variations in the horizontal and vertical directions near the corners of the patch clearly manifests as structured noise in the simulation.
A simulated patch of the same size as the training patches ($64\times64$) is shown in Figure~\ref{fig:bg_ripple_64}.
The patch shows a noisy texture covering its whole area, which is originated from the complete superposition of the \emph{ripples} in the $x$- and $y$-axis. 
The generated noise pattern is similar to the noise observed in the target photographs.

For further analysis of such effect, we compare the noise profiles of the original simulation targets with those produced by our best-performing models trained with L2-norm, cross-entropy, and LPIPS.
For brevity, we omit the model trained with focal loss, as the results for this training objective are, for all practical purposes, indistinguishable from those of the cross-entropy-based models.
This comparison is conducted on a dataset from a real wafer.

\begin{figure*}[ht]
\captionsetup{justification=centering}
\centering
\begin{subfigure}[t]{0.2\linewidth}
  \centering
  \includegraphics[width=0.98\linewidth]{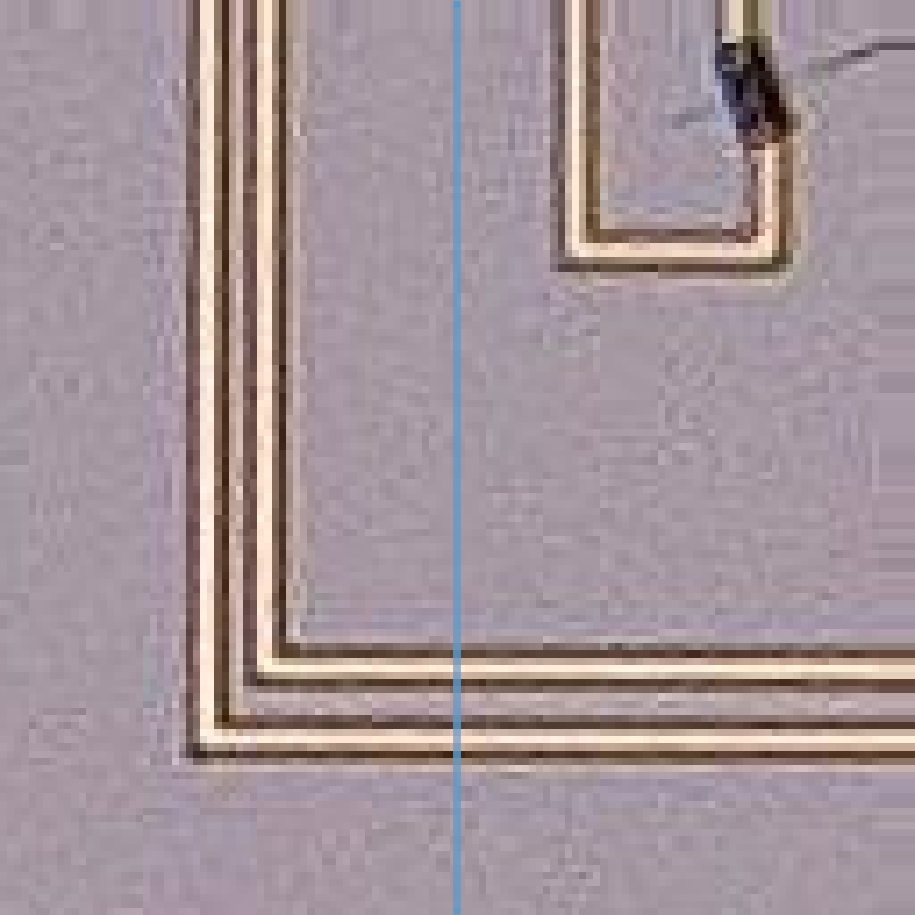}
  \caption{Target photo}
  \label{fig:ripple_sample_ph}
\end{subfigure}%
\begin{subfigure}[t]{0.2\linewidth}
  \centering
  \includegraphics[width=0.98\linewidth]{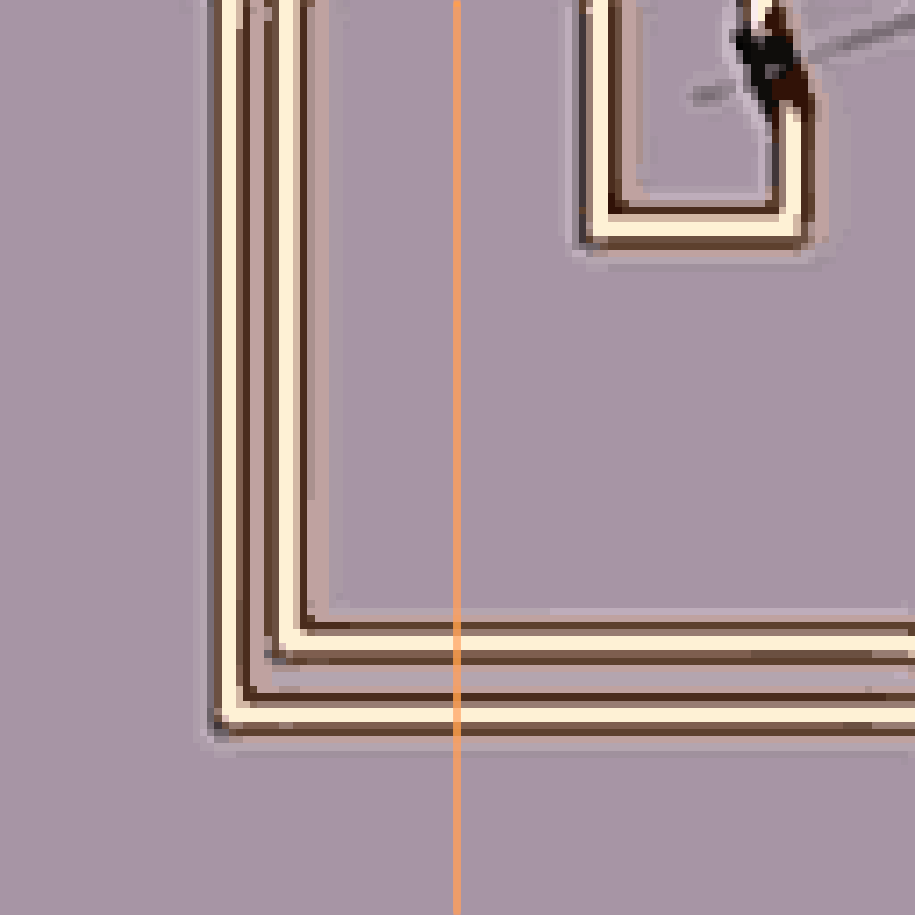}
  \caption{Cross-entropy}
  \label{fig:ripple_sample_xe}
\end{subfigure}%
\begin{subfigure}[t]{0.2\linewidth}
  \centering
  \includegraphics[width=0.98\linewidth]{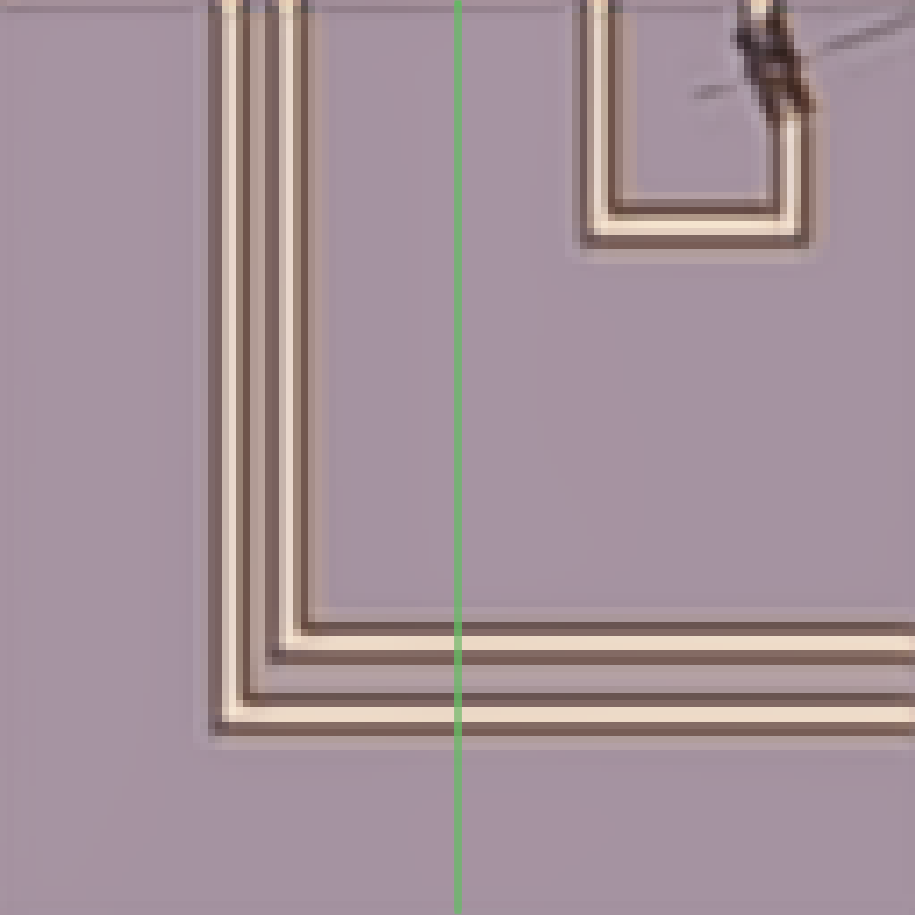}
  \caption{L2}
  \label{fig:ripple_sample_l2}
\end{subfigure}%
\begin{subfigure}[t]{0.2\linewidth}
  \centering
  \includegraphics[width=0.98\linewidth]{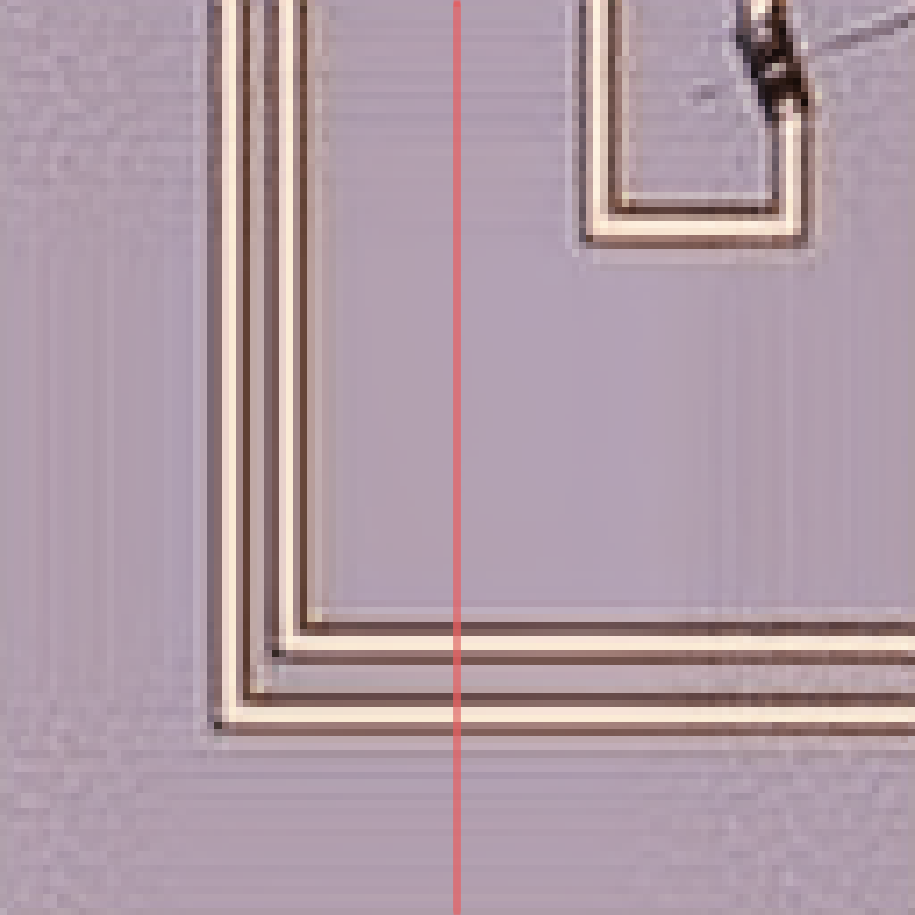}
  \caption{LPIPS}
  \label{fig:ripple_sample_lpips}
\end{subfigure}%
\captionsetup{justification=justified}
\caption{A target patch and simulation samples, all 128$\times$128-pixels in size.
The original photo contains noise, which is generally absent from the semantic segmentation-based simulation (in~\ref{fig:ripple_sample_xe}, with reconstruction L2 = 0.034) and from the L2-based simulation (in~\ref{fig:ripple_sample_l2}, with L2 = 0.023).
The LPIPS based simulation in~\ref{fig:ripple_sample_lpips} has high computed similarity to the target photo (L2 = 0.030), while also presenting visible artifacts around its periphery.
The highlighted columns in the center of each patch correspond to the respective path used to plot the color profiles in Figures~\ref{fig:ripple_plot} and~\ref{fig:detail_ripple_plot}.
}
\label{fig:ripple_sample}
\end{figure*}

\begin{figure}[ht]
\centering
\includegraphics[width=0.76\linewidth]{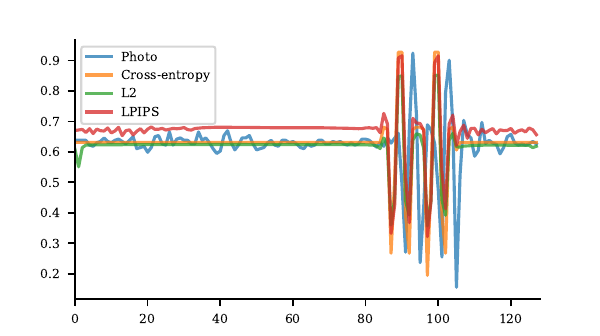}
\caption{Color profile along a vertical path through the centerline of simulated patches from Figure\ref{fig:ripple_sample}.
The patches are square and 128-pixel wide, with coordinates increasing in the $y$-axis from top to bottom.
Values on the RGB channel were averaged, to obtain a single scalar per pixel in the column.
Notice that both the models trained on cross-entropy and on L2-norm as loss functions show a smooth color profile on the background areas.
The model trained on focal loss is omitted for brevity, but exhibits the same averaging effect.
The model trained on LPIPS, however, shows a rugged profile which, although differing in mean-value, better mimics the original photo's noise profile.
It is also noteworthy to mention the misalignment effect between ground-truth and CAD layers, as the original photo (\(\color[rgb]{0.341, 0.6, 0.776}\bullet\)) shows peaks in different locations }
\label{fig:ripple_plot}
\end{figure}

\begin{figure}[ht]
\centering
\includegraphics[width=0.6\linewidth]{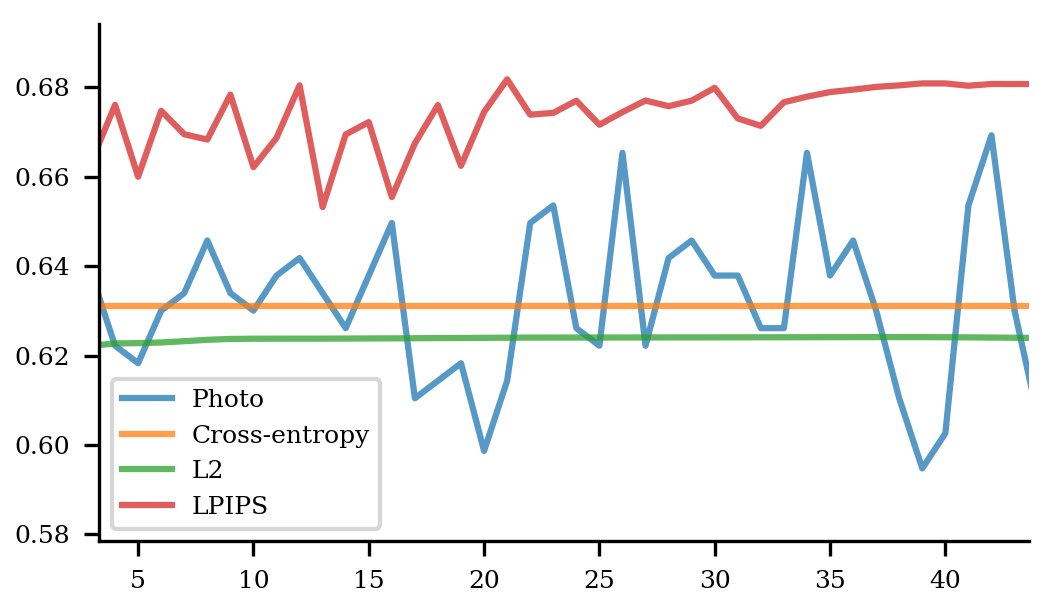}
\caption{Detailed view of the same plot as in Figure~\ref{fig:ripple_plot}, focusing on the area containing the undulating artifact in the top section of the simulated sample.
It is evident that the model trained on LPIPS simulates noise better than the remaining portrayed models, showing a similar characteristic to the profile on the original photo.
From the 32-nd row onward, the effect diminishes, as also seen in Figure~\ref{fig:bg_ripple_128}.}
\label{fig:detail_ripple_plot}
\end{figure}

We simulate $128\times128$-pixel patches and analyze the noise profile along the image's center column.
The simulations presented in Figure~\ref{fig:ripple_sample} are used without loss of generality with other datasets.
To simplify the comparison, RGB values are averaged into a single scalar per pixel. The resulting profiles are shown in Figures~\ref{fig:ripple_plot} and~\ref{fig:detail_ripple_plot}.

The color profiles shown in Figure~\ref{fig:ripple_plot} suggest that training with LPIPS as a loss function improves the simulation of noise.
This effect may be attributed to the LPIPS objective encouraging the neural network to leverage its downsampling and upsampling pathways, resulting in more structured and realistic noise characteristics.

Overall, our results demonstrate that both decision trees and neural networks are generally effective at simulating defect-free wafers from defective source images, despite some susceptibility to training instability.
Semantic segmentation models applied to quantized images outperform regression models operating in RGB space.
Decision trees, used as a naïve benchmark, are also a potentially viable method but perform less favorably.
The ripple artifacts discussed earlier highlight the nuanced behavior of these models under different training conditions and loss functions.

\section{Towards defect detection}
\label{defection}

According to \cite{wittmann2022photorealistic}, the manual inspection of an entire wafer requires approximately 20 hours of work by qualified personnel—representing nearly the half of a workweek that could otherwise be allocated to higher-value tasks.
In this context, the proposed simulation method holds practical relevance if used as a means of enhancing productivity within any specific stages of the foundry workflow.
In this section, we demonstrate an automated surface defect detection methodology based on the golden die simulation as one potential way to offer tangible benefits in MPW manufacturing.

Firstly, to apply the simulation method presented above to surface defect detection via template matching, it is crucial to assess that the simulated wafer effectively represents a golden die, generating perceptually realistic images that do not contain the defects present in the training images.
Finally, it would be beneficial that the score maps generated for defect detection are properly evaluated, in order to assert the generality of our method.

\begin{figure}[ht]
\captionsetup{justification=centering}
\centering
\begin{subfigure}[t]{0.25\linewidth}
  \centering
  \includegraphics[width=0.92\linewidth]{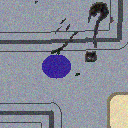}
  \caption{Target sample.}
  \label{fig:defection_toy1}
\end{subfigure}%
\begin{subfigure}[t]{0.25\linewidth}
  \centering
  \includegraphics[width=0.92\linewidth]{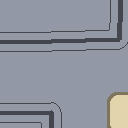}
  \caption{Simulation.}
  \label{fig:defection_toy2}
\end{subfigure}%
\begin{subfigure}[t]{0.25\linewidth}
  \centering
  \includegraphics[width=0.92\linewidth]{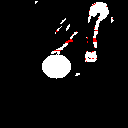}
  \caption{Binarized defect map.}
  \label{fig:defection_toy3}
\end{subfigure}%
\captionsetup{justification=justified}
\caption{The defect detection works considerably well on the synthetic datasets, as we have complete control of the procedural generation of defects.
The average precision for this patch is 0.98.
It shows the results of a regression model trained with L2 loss.}
\label{fig:defection_toy}
\end{figure}

Measuring the pixelwise dissimilarity between a simulation and the corresponding photographic sample on the wafer, we can obtain a dissimilarity score map.
In Figure~\ref{fig:defection_toy} we can observe an example of a defect detector generated by obtaining a score map with pixelwise mean squared error --- the L2-norm --- as a similarity metric.

Since we have access to perfect pixel-level annotations for defects on the synthetic wafers, we are able to quantitatively evaluate the performance of this detector.
Notably the dataset in this setting is heavily unbalanced, since the predicted negative samples -- i.e. non-defects -- outnumber the predicted positive samples by several orders of magnitude.
To address this characteristic we choose to evaluate the score map by producing its \emph{average precision}, also called area under the precision-recall curve, which does not account for the true negatives in the predicted sample.
The automated defect detection by template matching in synthetic data is virtually perfect when the simulation is good enough.
Average precision is computed with a variable threshold on the pixelwise L2 scores.
For ease of visualization, we show a binarized version of the score map in Figure~\ref{fig:defection_toy3}, using a constant threshold value of 0.1 for detection.
As demonstrated in Section~\ref{simulation}, this threshold exceeds the typical L2-norm values observed in successful simulations of perceptually similar samples.

\begin{figure}[ht]
\captionsetup{justification=centering}
\centering
\begin{subfigure}[t]{0.25\linewidth}
  \centering
  \includegraphics[width=0.92\linewidth]{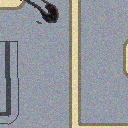}
  \caption{Target sample.}
  \label{fig:defection_hallu1}
\end{subfigure}%
\begin{subfigure}[t]{0.25\linewidth}
  \centering
  \includegraphics[width=0.92\linewidth]{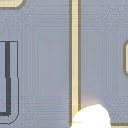}
  \caption{Simulation.}
  \label{fig:defection_hallu2}
\end{subfigure}%
\begin{subfigure}[t]{0.25\linewidth}
  \centering
  \includegraphics[width=0.92\linewidth]{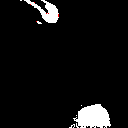}
  \caption{Binarized defect map.}
  \label{fig:defection_hallu3}
\end{subfigure}%
\captionsetup{justification=justified}
\caption{The defect detection fails completely when there is an anomaly in the simulated patch.
Average precision on this patch is 0.44. Regression model trained with LPIPS loss on synthetic data.}
\label{fig:defection_hallu}
\end{figure}

Since this method assumes a perfectly simulated, defect-free sample, it fails when a simulation contains hallucinations.
In Figure~\ref{fig:defection_hallu} we can note that the defect detection fails when the dissimilarity stems from the simulation process rather than the manufacturing process.
All defects in the sample from Figure~\ref{fig:defection_hallu1} were adequately detected.
However, the hallucinations from Figure~\ref{fig:defection_hallu2} logically reflect on the score map and are picked as false positives in Figure~\ref{fig:defection_hallu3}.
As a result, the computed average precision in this case is unreliable, as it is affected by artifacts not corresponding to actual defects.

\begin{figure}[ht]
\captionsetup{justification=centering}
\centering
\begin{subfigure}[t]{0.25\linewidth}
  \centering
  \includegraphics[width=0.92\linewidth]{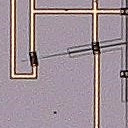}
  \caption{Target sample.}
  \label{fig:defection_rl1}
\end{subfigure}%
\begin{subfigure}[t]{0.25\linewidth}
  \centering
  \includegraphics[width=0.92\linewidth]{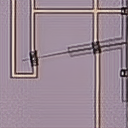}
  \caption{Simulation.}
  \label{fig:defection_rl2}
\end{subfigure}%
\begin{subfigure}[t]{0.25\linewidth}
  \centering
  \includegraphics[width=0.92\linewidth]{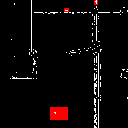}
  \caption{Binarized defect map.}
  \label{fig:defection_rl3}
\end{subfigure}%
\captionsetup{justification=justified}
\caption{Regression model trained with LPIPS loss on a real wafer.
Due to the stitching process in the microscope, the photo might be not properly aligned with the CAD layers.
This causes thin structures to show in the score map as dissimilarities.
Post simulation realignment might be necessary to accurately catch defects.
The labeled areas, in red, are also bigger than the defect instances.
Those factors turn the pixelwise evaluation inaccurate.}
\label{fig:defection_rl}
\end{figure}

On Figure~\ref{fig:defection_rl} we show the application of the same detection and evaluation method presented above, but this time applied to a real wafer.
On real data we face two main challenges for the application of the proposed method.
Firstly, we do not have access to such detailed labels as in the synthetic dataset, as many defects were labeled as a rectangular bounding box exceeding the dimensions of the defect itself.
The red regions in Figure~\ref{fig:defection_rl3} show the available annotations.
The pixelwise evaluation presented is then inaccurate, since the labels effectively mark true negative predictions as false negatives, skewing the computation of recall.
Secondly, misalignment stemming from the stitching process of the photos and the homographic transformations to align the CAD layers generate simulations that, while well aligned to the CAD layouts, might be offset when superimposed to the wafer photos.
As a result, thin and elongated mismatched edges produce high dissimilarity scores, being detected as defects in the score map if directly compared.

\begin{figure}[ht]
\captionsetup{justification=centering}
\centering
\begin{subfigure}[t]{0.25\linewidth}
  \centering
  \includegraphics[width=0.92\linewidth]{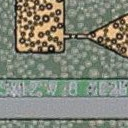}
  \caption{Target sample.}
  \label{fig:defection_hd1}
\end{subfigure}%
\begin{subfigure}[t]{0.25\linewidth}
  \centering
  \includegraphics[width=0.92\linewidth]{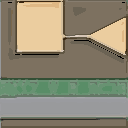}
  \caption{Simulation.}
  \label{fig:defection_hd2}
\end{subfigure}%
\begin{subfigure}[t]{0.25\linewidth}
  \centering
  \includegraphics[width=0.92\linewidth]{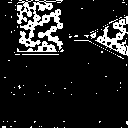}
  \caption{Score map.}
  \label{fig:defection_hd3}
\end{subfigure}%
\captionsetup{justification=justified}
\caption{Classification model trained with focal loss, on a wafer with very high defect density.
It is evident that the learned model successfully simulates a defect-free sample,
and that the pixelwise similarity metric reveals the position of most of the defects with significant performance.}
\label{fig:defection_hd}
\end{figure}

In Figure~\ref{fig:defection_hd} we see a predicted sample from a model trained on a wafer affected by unusually high defect density.
The simulation clearly depicts a defect-free version of the same wafer, and the dissimilarity score map is markedly effective in localizing the defects on the target sample.
Unfortunately, we are unable to quantify these claims with the computation of the average precision due to the lack of labeled data for the wafer in question.

The choice of similarity metric for generating a score map largely depends on the data format.
In principle, any similarity or distance function may be employed, though some may be more appropriate than others depending on the task.
Pixelwise metrics are typically faster to compute in modern GPU accelerated hardware.
They can, however, be sensitive to noise and misalignment, potentially leading to unreliable detections.
Alternatively, metrics that yield patchwise scalar values can be adapted to provide a single value for each pixel by averaging over a sliding window which applies the desired metric successively over overlapping patches.
This process incurs in additional computational cost, but local deviations get averaged for a smoother result.
Perceptual metrics such as LPIPS, for example, can be applied in this manner to provide spatially resolved dissimilarity maps.
In addition, classification models may be evaluated in higher dimensional logit space, where semantic segmentation metrics can incorporate class confidence scores, rather than relying solely on the lower dimensional final predicted labels.

\section{Discussion}
\label{discussion}

In the proposed method, defect detection and training of simulation models are closely linked, since the functions used as training targets can be simultaneously used as similarity metrics to evaluate simulation quality and to generate a defect score map.
Considering that CAD layers and the photographic representation of the wafers are intrinsic to the manufacturing process, we define our method with the term \emph{unsupervised} in the sense that we do not use \emph{labeled defect data} to achieve defect-free simulations.
The quantitative evaluation of unsupervised methods without any ground-truth data to compare against is only possible to a very limited extent.
In the simulation setting, defective data is present in the dataset but \emph{a priori} not labeled, so at first view indistinguishable from good samples.
Therefore, even though similarity metrics can be applied to measure simulation quality, reported scores will not be absolute, as the evaluation target contains defects which shouldn't be learned at all.

Not only our simulation target contains defects, it also contains noise.
Our experiments show that models trained on different objective functions learn differently how to deal with noise, and such differences are not accurately reflected in patch-averaged metrics.
While some models adopt an \emph{averaging} approach, effectively removing noise from simulations, others simulate realistic looking noise in the final simulation, without noticeable effects in the computed similarity  metrics.
The desired appearance of a simulation might depend on the application context.
For employment in defect detection, for example, there is a greater interest in the actual detection performance than necessarily in perceptual acuity.

Following the simulation step, a score map for defect detection can be generated using the same similarity metrics.
The major advantage of the proposed method is that it does not need any labeling for training and generation, realizing the goal of minimizing human labor.
The evaluation of this score map poses additional challenges, however, as its focus is not on enhancing average quality at the patch level, but rather on achieving precise pixel-wise detections.
Such an evaluation would require properly labeled defect data.
While we do have sections of the wafer which were completely labeled by qualified personell, poor label quality also limits the evaluation's accuracy.

For the reasons mentioned above, a fully automated unsupervised defect detection pipeline remains mostly as an open challenge, even with a reliable golden standard simulator.
It is not hard to envisage that derivations of this work could also serve as a stepping stone toward \emph{human-in-the-loop} applications.
The dissimilarity maps can be used, for example, to guide the annotation process, isolating probable defects within highly unbalanced datasets, cornering the \emph{needle-in-a-haystack} aspect of this problem.
A small number of correctly labeled detections can be used to further train defect classification networks.

In semiconductor production, MPW runs are both time-consuming and costly, with multiple runs often being simultaneously processed at different stages of fabrication.
Since the CAD layers used represent steps of the process, this simulation method could also be used in intermediate steps, for early defect detection in production.
Each MPW run takes several months from beginning to finish, whereas the models showed in this work can be trained on modern hardware within a span of hours.
The difference in process time-scale between neural network training and actual manufacturing suggests that the proposed pipeline shows potential to deployment in less powerful, more energy-efficient computers, as well as a possible pathway for integration in embedded hardware in the foundry.

\section{Conclusion}
\label{conc}

In this study, we have demonstrated the feasibility and effectiveness of simulating realistic images of photonic wafers at the chip-level using a convolutional neural network, formulating the problem both as a semantic segmentation problem and a regression task.
Notably, even when utilizing defective wafers as simulation targets, the trained models yield defect-free simulations, and a small portion of the wafer proving sufficient for generalization across the entire wafer.
The generated images can in turn be used as \emph{golden standard} wafers for template matching-based defect detection.

Besides the aforementioned general simulation method, our contribution also encompasses a benchmarking for evaluation of such simulations, 
the notion that a learned perceptual similarity metric is suitable as a loss function for training --- although different training objectives can be used for their own characteristic behaviors ---
and a clear pathway to automated chip-level defect detection on low-scale multi-project photonic InP wafers.
Additionally, we present tools for generating synthetic wafer datasets that can facilitate further advancement of the described techniques.

In general, the scarcity of datasets in this domain limits the evaluation of our simulation results and the quality of the trained models.
Furthermore, the evaluation of defect detection performance is restricted by the absence of well segmented defect labels.
Future efforts aimed at creating open and comprehensive datasets would highly benefit the field.
Additional work should also focus on data pre-processing,
as the high sensitivity of the detectors may lead to misclassification of noise as false positives or ignore small defects as false negatives.
Employing higher-resolution, low-noise images would be an approachable way to mitigate this classic trade-off.
The methods would also benefit from improved alignment between the CAD layers and the photographic images, and post-inference greater alignment between simulatioons and wafer photographs.
Other future avenues of research could focus on optimizing different components within the pipeline, including
the selection of hyperparameters, target functions and pre- and post-processing heuristics.
Despite these limitations, our work illustrates the substantial potential of applying semantic segmentation-based simulation for surface defect detection applications in the MPW InP production line.

\bibliographystyle{elsarticle-harv}
\bibliography{bib}

\newpage
\appendix

\section{Evaluation metrics and complete quantitative similarity results}
\label{ap:results}

We perform quantitative evaluation of simulation quality by computation of a few selected similarity metrics between simulation target (photograph of the wafer) and the simulated patch.
Besides metrics commonly found in the literature, we extend the 1-off accuracy \cite{eidinger2014age} to $k$-off accuracy, by considering a prediction correct if the target color is situated in one of the $k$ adjacent bins to the predicted bin in the adopted palette.
Here we report $k$-off for $k=2$. 
L2, LPIPS and PSNR were applied to the RGB images.
Cross-entropy and $k$-off accuracy were applied to the quantized photographs and the prediction logits, and thus are not available for models trained on the regression task.
A few datasets (R6 and R7) do not contain RGB images, and were therefore only trained on the semantic segmentation/classification task.

PSNR, LPIPS and $k$-off accuracy implementations return a scalar value for a whole patch.
L2 returns the averaged value of a pixelwise computation, as does cross-entropy.
The patch-wise results for all patches in the validation split were averaged, and we report mean value and standard deviation among patches in the tables that follow.
Table~\ref{tab:results_real} shows the computed metrics for models trained on real data and Table~\ref{tab:results_mock} shows the same set of metrics applied for models trained on synthethic data.
On Table~\ref{tab:results_tree} we report the obtained similarity metrics for the decision trained to simulate both real wafer datasets and synthetic datasets.
We report the 10-th epoch result for all models, and the best performing model in regards to both LPIPS and L2.
Some specific checkpoints were the best-performing of a run for both metrics simultaneously, and in a few cases the model at the 10-th epoch was also the best-performing checkpoint.
In particular, we can also observer how $k$-off accuracy is low for models from Table~\ref{tab:results_mock}, even though the models overall performance has been shown to be acceptable.
This effect originates from the sub-optimal ordering of colors in the synthetic pallette, as shown in Figure~\ref{fig:palette_mock}.
The $k$-off accuracy was therefore omitted from the main text of this report.

\begin{landscape}
\begin{longtable}{cccrlrlrlrlrl}
\caption{Quantitative results for synthetic wafers. Result shown in \emph{italic} when evaluation metric is the same as the training target.} \label{tab:results_real} \\
\toprule
Dataset & Training & Epoch & \multicolumn{2}{c}{L2 $\downarrow$} & \multicolumn{2}{c}{LPIPS $\downarrow$} & \multicolumn{2}{c}{PSNR $\uparrow$} & \multicolumn{2}{c}{Cross entropy $\downarrow$} & \multicolumn{2}{c}{$k$-off $\uparrow$} \\
    & objective  &  & mean & (SD) & mean & (SD) & mean & (SD) & mean & (SD) & mean & (SD) \\
\midrule
\endfirsthead
\caption[]{Quantitative results for synthetic wafers. Result shown in \emph{italic} when evaluation metric is the same as the training target.} \\
\toprule
Dataset & Training & Epoch & \multicolumn{2}{c}{L2 $\downarrow$} & \multicolumn{2}{c}{LPIPS $\downarrow$} & \multicolumn{2}{c}{PSNR $\uparrow$} & \multicolumn{2}{c}{Cross entropy $\downarrow$} & \multicolumn{2}{c}{$k$-off $\uparrow$} \\
    & objective  &  & mean & (SD) & mean & (SD) & mean & (SD) & mean & (SD) & mean & (SD) \\
\midrule
\endhead
\midrule
\multicolumn{13}{r}{Continued on next page} \\
\midrule
\endfoot
\bottomrule
\endlastfoot
R1 & Focal & 10 & 0.006 & (0.009) & 0.541 & (0.152) & 25.715 & (5.308) & 2.559 & (0.491) & 0.558 & (0.169) \\
R1 & Focal & 9 & 0.006 & (0.009) & 0.546 & (0.154) & 25.757 & (5.455) & 2.554 & (0.474) & 0.602 & (0.150) \\
R1 & LPIPS & 10 & 0.009 & (0.010) & \emph{0.544} & (0.144) & 23.135 & (5.487) & -- & -- & -- & -- \\
R1 & LPIPS & 2 & 0.007 & (0.010) & \emph{0.518} & (0.156) & 24.703 & (4.761) & -- & -- & -- & -- \\
R1 & L2 & 10 & \emph{0.005} & (0.007) & 0.579 & (0.140) & 24.385 & (3.708) & -- & -- & -- & -- \\
R1 & L2 & 7 & \emph{0.005} & (0.008) & 0.562 & (0.137) & 25.908 & (4.904) & -- & -- & -- & -- \\
R1 & L2 & 1 & \emph{0.006} & (0.008) & 0.571 & (0.138) & 24.660 & (4.387) & -- & -- & -- & -- \\
R1 & Cr. entropy & 10 & 0.005 & (0.008) & 0.543 & (0.149) & 26.388 & (5.556) & \emph{2.352} & (0.447) & 0.673 & (0.146) \\
R1 & Cr. entropy & 9 & 0.005 & (0.008) & 0.543 & (0.149) & 26.001 & (5.368) & \emph{2.332} & (0.397) & 0.664 & (0.131) \\
R2 & Focal & 10 & 0.006 & (0.009) & 0.544 & (0.177) & 24.907 & (5.057) & 2.703 & (0.862) & 0.593 & (0.121) \\
R2 & Focal & 4 & 0.008 & (0.012) & 0.548 & (0.165) & 24.315 & (5.554) & 2.998 & (0.876) & 0.590 & (0.121) \\
R2 & LPIPS & 10 & 0.011 & (0.015) & \emph{0.529} & (0.147) & 23.138 & (5.520) & -- & -- & -- & -- \\
R2 & LPIPS & 1 & 0.007 & (0.014) & \emph{0.521} & (0.156) & 24.644 & (5.166) & -- & -- & -- & -- \\
R2 & LPIPS & 4 & 0.006 & (0.009) & \emph{0.518} & (0.162) & 25.056 & (5.183) & -- & -- & -- & -- \\
R2 & L2 & 10 & \emph{0.006} & (0.007) & 0.568 & (0.158) & 24.513 & (4.213) & -- & -- & -- & -- \\
R2 & L2 & 8 & \emph{0.006} & (0.009) & 0.552 & (0.160) & 25.018 & (5.220) & -- & -- & -- & -- \\
R2 & L2 & 4 & \emph{0.006} & (0.008) & 0.560 & (0.149) & 25.632 & (5.247) & -- & -- & -- & -- \\
R2 & Cr. entropy & 10 & 0.008 & (0.012) & 0.553 & (0.175) & 23.868 & (5.613) & \emph{2.778} & (1.142) & 0.537 & (0.179) \\
R2 & Cr. entropy & 4 & 0.008 & (0.011) & 0.548 & (0.170) & 24.444 & (5.474) & \emph{3.913} & (3.173) & 0.591 & (0.121) \\
R2 & Cr. entropy & 3 & 0.009 & (0.022) & 0.550 & (0.171) & 25.221 & (6.321) & \emph{2.716} & (1.522) & 0.647 & (0.175) \\
R3 & Focal & 10 & 0.007 & (0.012) & 0.498 & (0.137) & 26.957 & (7.177) & 2.367 & (0.715) & 0.559 & (0.172) \\
R3 & Focal & 6 & 0.007 & (0.012) & 0.495 & (0.141) & 27.120 & (7.167) & 2.369 & (0.817) & 0.560 & (0.173) \\
R3 & LPIPS & 10 & 0.006 & (0.009) & \emph{0.477} & (0.139) & 26.265 & (5.659) & -- & -- & -- & -- \\
R3 & LPIPS & 20 & 0.007 & (0.009) & \emph{0.483} & (0.136) & 25.780 & (6.235) & -- & -- & -- & -- \\
R3 & LPIPS & 6 & 0.007 & (0.009) & \emph{0.480} & (0.132) & 25.876 & (6.589) & -- & -- & -- & -- \\
R3 & LPIPS & 9 & 0.006 & (0.009) & \emph{0.470} & (0.136) & 26.550 & (5.987) & -- & -- & -- & -- \\
R3 & L2 & 10 & \emph{0.006} & (0.009) & 0.509 & (0.132) & 26.955 & (6.540) & -- & -- & -- & -- \\
R3 & L2 & 9 & \emph{0.006} & (0.009) & 0.507 & (0.131) & 27.182 & (6.684) & -- & -- & -- & -- \\
R3 & L2 & 2 & \emph{0.006} & (0.008) & 0.519 & (0.122) & 26.895 & (6.539) & -- & -- & -- & -- \\
R3 & Cr. entropy & 10 & 0.007 & (0.011) & 0.496 & (0.141) & 26.629 & (6.736) & \emph{2.419} & (0.857) & 0.454 & (0.190) \\
R3 & Cr. entropy & 5 & 0.007 & (0.011) & 0.500 & (0.137) & 26.032 & (6.462) & \emph{2.668} & (1.682) & 0.453 & (0.192) \\
R4 & Focal & 10 & 0.007 & (0.013) & 0.533 & (0.173) & 25.836 & (6.211) & 2.517 & (0.864) & 0.615 & (0.153) \\
R4 & Focal & 9 & 0.006 & (0.013) & 0.526 & (0.173) & 26.196 & (5.956) & 2.469 & (0.779) & 0.603 & (0.151) \\
R4 & Focal & 4 & 0.007 & (0.013) & 0.532 & (0.171) & 25.755 & (6.229) & 2.519 & (0.876) & 0.597 & (0.156) \\
R4 & LPIPS & 10 & 0.006 & (0.010) & \emph{0.527} & (0.162) & 24.536 & (4.165) & -- & -- & -- & -- \\
R4 & LPIPS & 8 & 0.006 & (0.013) & \emph{0.485} & (0.162) & 25.704 & (5.186) & -- & -- & -- & -- \\
R4 & LPIPS & 7 & 0.005 & (0.012) & \emph{0.481} & (0.155) & 25.632 & (4.800) & -- & -- & -- & -- \\
R4 & L2 & 10 & \emph{0.010} & (0.017) & 0.565 & (0.166) & 24.647 & (6.519) & -- & -- & -- & -- \\
R4 & L2 & 78 & \emph{0.005} & (0.011) & 0.539 & (0.166) & 26.443 & (5.203) & -- & -- & -- & -- \\
R4 & L2 & 40 & \emph{0.006} & (0.013) & 0.532 & (0.166) & 26.125 & (5.391) & -- & -- & -- & -- \\
R4 & Cr. entropy & 10 & 0.007 & (0.014) & 0.533 & (0.179) & 26.038 & (5.835) & \emph{2.567} & (1.200) & 0.590 & (0.146) \\
R4 & Cr. entropy & 65 & 0.006 & (0.013) & 0.529 & (0.184) & 26.595 & (5.953) & \emph{2.542} & (1.102) & 0.599 & (0.155) \\
R4 & Cr. entropy & 6 & 0.006 & (0.013) & 0.535 & (0.181) & 26.020 & (5.908) & \emph{2.669} & (1.723) & 0.595 & (0.146) \\
R5 & Focal & 10 & 0.009 & (0.021) & 0.521 & (0.180) & 25.541 & (6.158) & 2.580 & (1.526) & 0.653 & (0.126) \\
R5 & Focal & 4 & 0.010 & (0.021) & 0.525 & (0.178) & 25.454 & (6.289) & 2.708 & (1.549) & 0.674 & (0.127) \\
R5 & Focal & 6 & 0.010 & (0.021) & 0.521 & (0.179) & 25.493 & (6.386) & 2.527 & (1.074) & 0.664 & (0.132) \\
R5 & LPIPS & 10 & 0.012 & (0.014) & \emph{0.526} & (0.169) & 21.670 & (4.365) & -- & -- & -- & -- \\
R5 & LPIPS & 18 & 0.008 & (0.016) & \emph{0.495} & (0.173) & 25.088 & (5.651) & -- & -- & -- & -- \\
R5 & LPIPS & 4 & 0.007 & (0.013) & \emph{0.483} & (0.174) & 24.610 & (4.515) & -- & -- & -- & -- \\
R5 & L2 & 10 & \emph{0.008} & (0.019) & 0.530 & (0.175) & 25.182 & (5.146) & -- & -- & -- & -- \\
R5 & L2 & 4 & \emph{0.006} & (0.011) & 0.526 & (0.171) & 26.091 & (5.202) & -- & -- & -- & -- \\
R5 & L2 & 3 & \emph{0.008} & (0.017) & 0.529 & (0.171) & 26.016 & (5.776) & -- & -- & -- & -- \\
R5 & Cr. entropy & 10 & 0.010 & (0.022) & 0.523 & (0.181) & 25.524 & (6.219) & \emph{2.470} & (0.715) & 0.669 & (0.116) \\
R5 & Cr. entropy & 1 & 0.010 & (0.020) & 0.524 & (0.176) & 25.134 & (6.410) & \emph{2.689} & (1.064) & 0.674 & (0.135) \\
R6 & Focal & 10 & 0.003 & (0.003) & 0.397 & (0.159) & 27.462 & (4.124) & 5.447 & (1.886) & 0.626 & (0.289) \\
R6 & Focal & 5 & 0.002 & (0.003) & 0.363 & (0.165) & 30.575 & (6.389) & 11.805 & (5.695) & 0.629 & (0.290) \\
R6 & Cr. entropy & 10 & 0.011 & (0.003) & 0.494 & (0.131) & 19.582 & (0.945) & \emph{12.064} & (4.342) & 0.616 & (0.285) \\
R6 & Cr. entropy & 1 & 0.002 & (0.003) & 0.363 & (0.164) & 30.353 & (6.179) & \emph{41.346} & (12.008) & 0.628 & (0.290) \\
R7 & Focal & 10 & 0.002 & (0.002) & 0.336 & (0.114) & 28.721 & (2.684) & 3.811 & (1.561) & 0.786 & (0.209) \\
R7 & Focal & 2 & 0.001 & (0.002) & 0.294 & (0.119) & 33.803 & (5.492) & 24.054 & (9.240) & 0.793 & (0.210) \\
R7 & Cr. entropy & 10 & 0.001 & (0.002) & 0.294 & (0.119) & 33.803 & (5.492) & \emph{46.933} & (24.520) & 0.793 & (0.210) \\
R7 & Cr. entropy & 4 & 0.001 & (0.002) & 0.294 & (0.119) & 33.702 & (5.400) & \emph{23.606} & (12.730) & 0.793 & (0.210) \\
R8 & Focal & 10 & 0.006 & (0.009) & 0.533 & (0.194) & 25.678 & (5.496) & 2.423 & (0.537) & 0.599 & (0.144) \\
R8 & Focal & 9 & 0.006 & (0.008) & 0.530 & (0.194) & 25.846 & (5.616) & 2.349 & (0.482) & 0.598 & (0.146) \\
R8 & LPIPS & 10 & 0.007 & (0.010) & \emph{0.531} & (0.179) & 24.554 & (5.111) & -- & -- & -- & -- \\
R8 & LPIPS & 6 & 0.006 & (0.007) & \emph{0.527} & (0.187) & 24.659 & (4.411) & -- & -- & -- & -- \\
R8 & LPIPS & 7 & 0.007 & (0.010) & \emph{0.527} & (0.175) & 24.507 & (4.985) & -- & -- & -- & -- \\
R8 & L2 & 10 & \emph{0.006} & (0.008) & 0.566 & (0.171) & 24.373 & (4.709) & -- & -- & -- & -- \\
R8 & L2 & 3 & \emph{0.006} & (0.008) & 0.566 & (0.165) & 25.185 & (4.985) & -- & -- & -- & -- \\
R8 & Cr. entropy & 10 & 0.006 & (0.008) & 0.538 & (0.198) & 24.652 & (4.555) & \emph{2.386} & (0.462) & 0.472 & (0.136) \\
R8 & Cr. entropy & 4 & 0.006 & (0.009) & 0.537 & (0.191) & 25.572 & (5.536) & \emph{2.341} & (0.374) & 0.599 & (0.141) \\
R8 & Cr. entropy & 7 & 0.006 & (0.008) & 0.534 & (0.192) & 25.484 & (5.571) & \emph{2.448} & (0.441) & 0.593 & (0.147) \\
R9 & Focal & 10 & 0.006 & (0.011) & 0.535 & (0.187) & 25.608 & (5.660) & 2.492 & (0.527) & 0.597 & (0.162) \\
R9 & Focal & 9 & 0.007 & (0.011) & 0.533 & (0.188) & 25.572 & (5.694) & 2.497 & (0.538) & 0.587 & (0.154) \\
R9 & LPIPS & 10 & 0.007 & (0.010) & \emph{0.500} & (0.181) & 24.550 & (4.643) & -- & -- & -- & -- \\
R9 & LPIPS & 1 & 0.008 & (0.011) & \emph{0.547} & (0.169) & 24.120 & (5.632) & -- & -- & -- & -- \\
R9 & LPIPS & 4 & 0.007 & (0.011) & \emph{0.505} & (0.172) & 24.867 & (5.134) & -- & -- & -- & -- \\
R9 & L2 & 10 & \emph{0.006} & (0.008) & 0.562 & (0.168) & 25.045 & (5.539) & -- & -- & -- & -- \\
R9 & L2 & 9 & \emph{0.006} & (0.010) & 0.555 & (0.171) & 25.648 & (5.505) & -- & -- & -- & -- \\
R9 & Cr. entropy & 10 & 0.009 & (0.017) & 0.544 & (0.176) & 25.050 & (6.342) & \emph{2.486} & (0.636) & 0.593 & (0.171) \\
R9 & Cr. entropy & 8 & 0.007 & (0.012) & 0.534 & (0.186) & 25.447 & (5.841) & \emph{2.462} & (0.528) & 0.601 & (0.161) \\
\end{longtable}

\begin{longtable}{cccrlrlrlrlrl}
\caption{Quantitative results for synthetic wafers. Result shown in \emph{italic} when evaluation metric is the same as the training target.} \label{tab:results_mock} \\
\toprule
Dataset & Training & Epoch & \multicolumn{2}{c}{L2 $\downarrow$} & \multicolumn{2}{c}{LPIPS $\downarrow$} & \multicolumn{2}{c}{PSNR $\uparrow$} & \multicolumn{2}{c}{Cross entropy $\downarrow$} & \multicolumn{2}{c}{$k$-off $\uparrow$} \\
    & objective  &  & mean & (SD) & mean & (SD) & mean & (SD) & mean & (SD) & mean & (SD) \\
\midrule
\endfirsthead
\caption[]{Quantitative results for synthetic wafers. Result shown in \emph{italic} when evaluation metric is the same as the training target.} \\
\toprule
Dataset & Training & Epoch & \multicolumn{2}{c}{L2 $\downarrow$} & \multicolumn{2}{c}{LPIPS $\downarrow$} & \multicolumn{2}{c}{PSNR $\uparrow$} & \multicolumn{2}{c}{Cross entropy $\downarrow$} & \multicolumn{2}{c}{$k$-off $\uparrow$} \\
    & objective  &  & mean & (SD) & mean & (SD) & mean & (SD) & mean & (SD) & mean & (SD) \\
\midrule
\endhead
\midrule
\multicolumn{13}{r}{Continued on next page} \\
\midrule
\endfoot
\bottomrule
\endlastfoot
    S1 & Focal & 10 & 0.004 & (0.002) & 0.459 & (0.105) & 24.056 & (1.712) & 3.476 & (0.114) & 0.213 & (0.029) \\
    S1 & Focal & 3  & 0.004 & (0.002) & 0.459 & (0.105) & 24.042 & (1.709) & 3.475 & (0.113) & 0.213 & (0.029) \\
    S1 & LPIPS & 10 & 0.005 & (0.002) & \emph{0.446} & (0.118) & 23.368 & (1.599) & -- & -- & -- & -- \\
    S1 & LPIPS & 17 & 0.005 & (0.002) & \emph{0.446} & (0.117) & 23.325 & (1.584) & -- & -- & -- & -- \\
    S1 & LPIPS & 8  & 0.005 & (0.002) & \emph{0.446} & (0.118) & 23.198 & (1.703) & -- & -- & -- & -- \\
    S1 & L2   & 10 & \emph{0.004} & (0.002) & 0.457 & (0.109) & 24.268 & (1.734) & -- & -- & -- & -- \\
    S1 & L2   & 22 & \emph{0.004} & (0.002) & 0.457 & (0.109) & 24.271 & (1.740) & -- & -- & -- & -- \\
    S1 & L2   & 2  & \emph{0.004} & (0.002) & 0.458 & (0.108) & 24.246 & (1.724) & -- & -- & -- & -- \\
    S1 & Cr. entropy   & 10 & 0.004 & (0.002) & 0.459 & (0.105) & 24.038 & (1.710) & \emph{3.471} & (0.110) & 0.213 & (0.029) \\
    S1 & Cr. entropy   & 5  & 0.004 & (0.002) & 0.459 & (0.105) & 24.047 & (1.710) & \emph{3.469} & (0.113) & 0.213 & (0.029) \\
    S1 & Cr. entropy   & 8  & 0.004 & (0.002) & 0.459 & (0.105) & 24.046 & (1.710) & \emph{3.469} & (0.113) & 0.213 & (0.029) \\
    S2 & Focal & 10 & 0.004 & (0.002) & 0.460 & (0.106) & 24.041 & (1.712) & 3.475 & (0.113) & 0.213 & (0.029) \\
    S2 & Focal & 8  & 0.004 & (0.002) & 0.460 & (0.106) & 24.040 & (1.712) & 3.475 & (0.113) & 0.213 & (0.029) \\
    S2 & Focal & 9  & 0.004 & (0.002) & 0.460 & (0.106) & 24.039 & (1.711) & 3.475 & (0.112) & 0.213 & (0.029) \\
    S2 & LPIPS & 10 & 0.005 & (0.002) & \emph{0.439} & (0.111) & 23.089 & (1.531) & -- & -- & -- & -- \\
    S2 & LPIPS & 3  & 0.005 & (0.002) & \emph{0.442} & (0.112) & 22.941 & (1.510) & -- & -- & -- & -- \\
    S2 & LPIPS & 28 & 0.005 & (0.002) & \emph{0.441} & (0.112) & 23.011 & (1.511) & -- & -- & -- & -- \\
    S2 & L2   & 10 & \emph{0.004} & (0.002) & 0.462 & (0.109) & 24.153 & (1.713) & -- & -- & -- & -- \\
    S2 & L2   & 5  & \emph{0.004} & (0.002) & 0.459 & (0.110) & 24.234 & (1.728) & -- & -- & -- & -- \\
    S2 & L2   & 4  & \emph{0.004} & (0.002) & 0.459 & (0.110) & 24.231 & (1.727) & -- & -- & -- & -- \\
    S2 & Cr. entropy   & 10 & 0.004 & (0.002) & 0.459 & (0.106) & 24.042 & (1.712) & \emph{3.469} & (0.112) & 0.213 & (0.029) \\
    S2 & Cr. entropy   & 2  & 0.004 & (0.002) & 0.461 & (0.106) & 24.017 & (1.713) & \emph{3.470} & (0.112) & 0.213 & (0.029) \\
    S2 & Cr. entropy   & 4  & 0.004 & (0.002) & 0.460 & (0.106) & 24.032 & (1.711) & \emph{3.469} & (0.111) & 0.213 & (0.029) \\
    S3 & Focal & 10 & 0.004 & (0.002) & 0.458 & (0.106) & 24.037 & (1.711) & 3.474 & (0.113) & 0.213 & (0.029) \\
    S3 & Focal & 9  & 0.004 & (0.002) & 0.458 & (0.106) & 24.034 & (1.710) & 3.475 & (0.114) & 0.213 & (0.029) \\
    S3 & Focal & 8  & 0.004 & (0.002) & 0.458 & (0.106) & 24.033 & (1.710) & 3.474 & (0.113) & 0.213 & (0.029) \\
    S3 & LPIPS & 10 & 0.005 & (0.002) & \emph{0.444} & (0.115) & 23.644 & (1.640) & -- & -- & -- & -- \\
    S3 & LPIPS & 8  & 0.005 & (0.002) & \emph{0.446} & (0.116) & 23.644 & (1.640) & -- & -- & -- & -- \\
    S3 & LPIPS & 1  & 0.005 & (0.002) & \emph{0.439} & (0.117) & 23.329 & (1.606) & -- & -- & -- & -- \\
    S3 & L2   & 10 & \emph{0.004} & (0.002) & 0.460 & (0.108) & 24.234 & (1.728) & -- & -- & -- & -- \\
    S3 & L2   & 9  & \emph{0.004} & (0.002) & 0.459 & (0.109) & 24.198 & (1.736) & -- & -- & -- & -- \\
    S3 & L2   & 7  & \emph{0.004} & (0.002) & 0.463 & (0.108) & 24.008 & (1.767) & -- & -- & -- & -- \\
    S3 & Cr. entropy   & 10 & 0.004 & (0.002) & 0.458 & (0.106) & 24.037 & (1.711) & \emph{3.467} & (0.112) & 0.213 & (0.029) \\
    S3 & Cr. entropy   & 4  & 0.004 & (0.002) & 0.458 & (0.106) & 24.024 & (1.710) & \emph{3.469} & (0.113) & 0.213 & (0.029) \\
    S3 & Cr. entropy   & 5  & 0.004 & (0.002) & 0.458 & (0.106) & 24.018 & (1.708) & \emph{3.469} & (0.113) & 0.213 & (0.029) \\
    S4 & Focal & 10 & 0.004 & (0.002) & 0.459 & (0.106) & 24.028 & (1.718) & 3.475 & (0.111) & 0.213 & (0.029) \\
    S4 & Focal & 5  & 0.004 & (0.002) & 0.459 & (0.106) & 24.026 & (1.722) & 3.475 & (0.110) & 0.213 & (0.029) \\
    S4 & LPIPS & 10 & 0.005 & (0.002) & \emph{0.441} & (0.113) & 23.335 & (1.589) & -- & -- & -- & -- \\
    S4 & LPIPS & 4  & 0.005 & (0.002) & \emph{0.454} & (0.119) & 23.566 & (1.619) & -- & -- & -- & -- \\
    S4 & LPIPS & 7  & 0.005 & (0.002) & \emph{0.443} & (0.117) & 23.061 & (1.542) & -- & -- & -- & -- \\
    S4 & L2   & 10 & \emph{0.005} & (0.002) & 0.475 & (0.102) & 23.502 & (1.803) & -- & -- & -- & -- \\
    S4 & L2   & 20 & \emph{0.004} & (0.002) & 0.459 & (0.109) & 24.212 & (1.736) & -- & -- & -- & -- \\
    S4 & L2   & 7  & \emph{0.005} & (0.002) & 0.471 & (0.106) & 23.660 & (1.853) & -- & -- & -- & -- \\
    S4 & L2   & 1  & \emph{0.005} & (0.003) & 0.466 & (0.107) & 23.505 & (1.956) & -- & -- & -- & -- \\
    S4 & Cr. entropy   & 10 & 0.004 & (0.002) & 0.459 & (0.106) & 24.032 & (1.722) & \emph{3.469} & (0.110) & 0.213 & (0.029) \\
    S4 & Cr. entropy   & 9  & 0.004 & (0.002) & 0.459 & (0.106) & 24.027 & (1.722) & \emph{3.469} & (0.110) & 0.213 & (0.029) \\
    S5 & Focal & 10 & 0.004 & (0.002) & 0.460 & (0.106) & 24.032 & (1.720) & 3.477 & (0.110) & 0.213 & (0.029) \\
    S5 & Focal & 5  & 0.004 & (0.002) & 0.460 & (0.106) & 24.025 & (1.720) & 3.473 & (0.111) & 0.213 & (0.029) \\
    S5 & LPIPS & 10 & 0.005 & (0.002) & \emph{0.453} & (0.118) & 22.996 & (1.533) & -- & -- & -- & -- \\
    S5 & LPIPS & 7  & 0.005 & (0.002) & \emph{0.453} & (0.118) & 23.011 & (1.570) & -- & -- & -- & -- \\
    S5 & LPIPS & 9  & 0.005 & (0.002) & \emph{0.452} & (0.118) & 23.018 & (1.553) & -- & -- & -- & -- \\
    S5 & L2   & 10 & \emph{0.004} & (0.002) & 0.459 & (0.110) & 24.231 & (1.741) & -- & -- & -- & -- \\
    S5 & L2   & 6  & \emph{0.004} & (0.002) & 0.459 & (0.109) & 24.231 & (1.736) & -- & -- & -- & -- \\
    S5 & L2   & 3  & \emph{0.004} & (0.002) & 0.462 & (0.109) & 24.129 & (1.708) & -- & -- & -- & -- \\
    S5 & Cr. entropy   & 10 & 0.004 & (0.002) & 0.460 & (0.106) & 24.032 & (1.721) & \emph{3.468} & (0.110) & 0.213 & (0.029) \\
    S5 & Cr. entropy   & 8  & 0.004 & (0.002) & 0.461 & (0.105) & 24.025 & (1.721) & \emph{3.476} & (0.109) & 0.213 & (0.029) \\
\end{longtable}

\begin{longtable}{ccrlrlrlrlrl}
\caption{Quantitative results for decision-tree based models. Quantitative results for synthetic wafers. Result shown in \emph{italic} when evaluation metric is the same as the training target.} \label{tab:results_tree} \\
\toprule
Dataset &  \multicolumn{2}{c}{L2 $\downarrow$} & \multicolumn{2}{c}{LPIPS $\downarrow$} & \multicolumn{2}{c}{PSNR $\uparrow$} & \multicolumn{2}{c}{Cross entropy $\downarrow$} & \multicolumn{2}{c}{$k$-off $\uparrow$} \\
      & mean & (SD) & mean & (SD) & mean & (SD) & mean & (SD) & mean & (SD) \\
\midrule
\endfirsthead
\caption[]{Quantitative results for synthetic wafers. Result shown in \emph{italic} when evaluation metric is the same as the training target.} \\
\toprule
Dataset & \multicolumn{2}{c}{L2 $\downarrow$} & \multicolumn{2}{c}{LPIPS $\downarrow$} & \multicolumn{2}{c}{PSNR $\uparrow$} & \multicolumn{2}{c}{Cross entropy $\downarrow$} & \multicolumn{2}{c}{$k$-off $\uparrow$} \\
        & mean & (SD) & mean & (SD) & mean & (SD) & mean & (SD) & mean & (SD) \\
\midrule
\endhead
\midrule
\multicolumn{11}{r}{Continued on next page} \\
\midrule
\endfoot
\bottomrule
\endlastfoot
R1 & 0.030 & (0.058) & 0.596 & (0.169) & 22.167 & (8.308) & 4.084 & (0.039) & 0.543 & (0.216) \\
R2 & 0.007 & (0.011) & 0.566 & (0.141) & 25.132 & (6.118) & 4.039 & (0.068) & 0.666 & (0.158) \\
R3 & 0.028 & (0.048) & 0.569 & (0.161) & 22.437 & (8.746) & 4.080 & (0.039) & 0.322 & (0.137) \\
R4 & 0.027 & (0.044) & 0.583 & (0.173) & 22.171 & (8.310) & 4.089 & (0.032) & 0.462 & (0.182) \\
R5 & 0.055 & (0.116) & 0.608 & (0.197) & 21.361 & (8.995) & 4.084 & (0.037) & 0.524 & (0.221) \\
R6 & 0.002 & (0.002) & 0.373 & (0.161) & 34.346 & (7.696) & 3.990 & (0.071) & 0.819 & (0.282) \\
R7 & 0.001 & (0.001) & 0.305 & (0.118) & 36.669 & (5.887) & 3.992 & (0.037) & 0.918 & (0.167) \\
R8 & 0.010 & (0.013) & 0.567 & (0.172) & 23.937 & (6.151) & 4.090 & (0.041) & 0.568 & (0.163) \\
R9 & 0.017 & (0.019) & 0.604 & (0.196) & 21.648 & (6.980) & 4.099 & (0.029) & 0.443 & (0.180) \\
S1 & 0.007 & (0.003) & 0.487 & (0.097) & 22.089 & (2.022) & 4.134 & (0.007) & 0.191 & (0.018) \\
S2 & 0.007 & (0.003) & 0.489 & (0.098) & 22.069 & (1.997) & 4.134 & (0.007) & 0.191 & (0.018) \\
S3 & 0.007 & (0.003) & 0.487 & (0.098) & 22.064 & (2.011) & 4.134 & (0.007) & 0.191 & (0.019) \\
S4 & 0.007 & (0.003) & 0.488 & (0.098) & 22.087 & (2.009) & 4.134 & (0.007) & 0.191 & (0.018) \\
S5 & 0.007 & (0.003) & 0.489 & (0.098) & 22.069 & (2.015) & 4.134 & (0.007) & 0.191 & (0.019) \\
\end{longtable}

\end{landscape}

To better contextualize the results, we also provide in tables~\ref{tab:corr_all}, \ref{tab:corr_tree}, \ref{tab:corr_mock} and~\ref{tab:corr_real} the agreement between measures of different similarity metrics applied to the models listed above.
The values above the main diagonal of the tables show the Pearson correlation coefficient, while the elements below the main diagonal show the Spearman's rank correlation.
In general, when analyzing Table~\ref{tab:corr_all}, the correlations follow the expected direction of alignment, but show considerable disagreement in several cases, highlighting the importance of careful consideration in the choice of evaluation metric for this particular setting.
More specifically, when computing the correlation coefficients separately for decision trees (Table~\ref{tab:corr_tree}, models trained on synthetic data (Table~\ref{tab:corr_mock}) and models trained on real wafers (Table~\ref{tab:corr_real}) we observe a substantial disagreement among a few metrics --- such as the seemingly uncorrelated nature of categorical cross entropy and focal loss to L1- and L2-norms for models which used synthetic datasets.

\begin{table}[ht]

\caption{Correlation coefficients between analyzed metrics, all models. Values above the main diagonal are the Pearson correlation coefficient, while the\colorbox{lightgray}{values below}the main diagonal correspond to the Spearman's rank correlation.}
\label{tab:corr_all}
\resizebox{\linewidth}{!}{%
\begin{tabular}{rcccccccccc}
\toprule
 & L1 $\downarrow$ & L2 $\downarrow$ & PSNR $\uparrow$ & SSIM $\uparrow$ & LPIPS $\downarrow$ & HaarPSI $\uparrow$ & $k$-off $\uparrow$ & Dice $\uparrow$ & Cr. entr. $\downarrow$ & Focal $\downarrow$ \\
\midrule
L1 $\downarrow$       & -- & 0.821 & -0.720 & -0.169 & 0.457 & -0.296 & -0.345 & -0.087 & -0.266 & -0.247 \\
    L2 $\downarrow$       & \cellcolor{lightgray} 0.255 & -- & -0.341 & 0.241 & 0.517 & -0.140 & 0.073 & -0.349 & -0.146 & -0.137 \\
    PSNR $\uparrow$       & \cellcolor{lightgray} -0.872 & \cellcolor{lightgray} 0.017 & -- & 0.455 & -0.436 & 0.512 & 0.644 & -0.351 & 0.508 & 0.491 \\
    SSIM $\uparrow$       & \cellcolor{lightgray} -0.587 & \cellcolor{lightgray} 0.446 & \cellcolor{lightgray} 0.789 & -- & 0.513 & 0.615 & 0.850 & -0.411 & -0.078 & -0.100 \\
    LPIPS $\downarrow$    & \cellcolor{lightgray} 0.033 & \cellcolor{lightgray} 0.737 & \cellcolor{lightgray} 0.212 & \cellcolor{lightgray} 0.561 & -- & 0.120 & 0.049 & 0.047 & -0.584 & -0.582 \\
    HaarPSI $\uparrow$    & \cellcolor{lightgray} -0.364 & \cellcolor{lightgray} 0.262 & \cellcolor{lightgray} 0.650 & \cellcolor{lightgray} 0.660 & \cellcolor{lightgray} 0.273 & -- & 0.548 & 0.019 & 0.072 & 0.047 \\
    $k$-off $\downarrow$  & \cellcolor{lightgray} -0.677 & \cellcolor{lightgray} 0.146 & \cellcolor{lightgray} 0.734 & \cellcolor{lightgray} 0.714 & \cellcolor{lightgray} 0.225 & \cellcolor{lightgray} 0.663 & -- & -0.639 & 0.248 & 0.228 \\
    Dice $\uparrow$       & \cellcolor{lightgray} 0.149 & \cellcolor{lightgray} -0.377 & \cellcolor{lightgray} -0.225 & \cellcolor{lightgray} -0.340 & \cellcolor{lightgray} -0.323 & \cellcolor{lightgray} -0.158 & \cellcolor{lightgray} -0.551 & -- & -0.410 & -0.413 \\
    Cr. entr. $\downarrow$& \cellcolor{lightgray} 0.347 & \cellcolor{lightgray} -0.286 & \cellcolor{lightgray} -0.409 & \cellcolor{lightgray} -0.595 & \cellcolor{lightgray} -0.425 & \cellcolor{lightgray} -0.571 & \cellcolor{lightgray} -0.272 & \cellcolor{lightgray} -0.293 & -- & 0.999 \\
    Focal $\downarrow$    & \cellcolor{lightgray} 0.356 & \cellcolor{lightgray} -0.287 & \cellcolor{lightgray} -0.417 & \cellcolor{lightgray} -0.603 & \cellcolor{lightgray} -0.417 & \cellcolor{lightgray} -0.570 & \cellcolor{lightgray} -0.278 & \cellcolor{lightgray} -0.301 & \cellcolor{lightgray} 0.998 & -- \\
\bottomrule
\end{tabular}

}%
\end{table}

\begin{table}[ht]
\caption{Correlation coefficients between analyzed metrics, trees. Values above the main diagonal are the Pearson correlation coefficient, while the\colorbox{lightgray}{values below}the main diagonal correspond to the Spearman's rank correlation.}
\label{tab:corr_tree}
\resizebox{\linewidth}{!}{%
\begin{tabular}{rcccccccccc}
\toprule
 & L1 $\downarrow$ & L2 $\downarrow$ & PSNR $\uparrow$ & SSIM $\uparrow$ & LPIPS $\downarrow$ & HaarPSI $\uparrow$ & $k$-off $\uparrow$ & Dice $\uparrow$ & Cr. entr. $\downarrow$ & Focal $\downarrow$ \\
\midrule
L1 $\downarrow$       & -- & 0.944 & -0.695 & 0.011 & 0.818 & -0.429 & -0.289 & -0.036 & 0.368 & 0.369 \\
L2 $\downarrow$       & \cellcolor{lightgray} 0.846 & -- & -0.454 & 0.236 & 0.689 & -0.158 & -0.011 & -0.281 & 0.092 & 0.092 \\
PSNR $\uparrow$       & \cellcolor{lightgray} -0.631 & \cellcolor{lightgray} -0.345 & -- & 0.585 & -0.818 & 0.918 & 0.795 & -0.587 & -0.867 & -0.867 \\
SSIM $\uparrow$       & \cellcolor{lightgray} -0.248 & \cellcolor{lightgray} 0.037 & \cellcolor{lightgray} 0.789 & -- & -0.039 & 0.829 & 0.884 & -0.990 & -0.871 & -0.871 \\
LPIPS $\downarrow$    & \cellcolor{lightgray} 0.780 & \cellcolor{lightgray} 0.934 & \cellcolor{lightgray} -0.420 & \cellcolor{lightgray} -0.020 & -- & -0.544 & -0.317 & 0.030 & 0.440 & 0.440 \\
HaarPSI $\uparrow$    & \cellcolor{lightgray} -0.367 & \cellcolor{lightgray} 0.007 & \cellcolor{lightgray} 0.732 & \cellcolor{lightgray} 0.873 & \cellcolor{lightgray} -0.011 & -- & 0.956 & -0.837 & -0.956 & -0.956 \\
    $k$-off $\downarrow$  & \cellcolor{lightgray} -0.314 & \cellcolor{lightgray} 0.059 & \cellcolor{lightgray} 0.666 & \cellcolor{lightgray} 0.846 & \cellcolor{lightgray} 0.007 & \cellcolor{lightgray} 0.969 & -- & -0.912 & -0.954 & -0.954 \\
    Dice $\uparrow$       & \cellcolor{lightgray} 0.138 & \cellcolor{lightgray} -0.125 & \cellcolor{lightgray} -0.582 & \cellcolor{lightgray} -0.855 & \cellcolor{lightgray} -0.125 & \cellcolor{lightgray} -0.811 & \cellcolor{lightgray} -0.833 & -- & 0.883 & 0.883 \\
    Cr. entr. $\downarrow$& \cellcolor{lightgray} 0.174 & \cellcolor{lightgray} -0.112 & \cellcolor{lightgray} -0.631 & \cellcolor{lightgray} -0.916 & \cellcolor{lightgray} -0.033 & \cellcolor{lightgray} -0.829 & \cellcolor{lightgray} -0.864 & \cellcolor{lightgray} 0.912 & -- & 1.000 \\
    Focal $\downarrow$    & \cellcolor{lightgray} 0.174 & \cellcolor{lightgray} -0.112 & \cellcolor{lightgray} -0.631 & \cellcolor{lightgray} -0.916 & \cellcolor{lightgray} -0.033 & \cellcolor{lightgray} -0.829 & \cellcolor{lightgray} -0.864 & \cellcolor{lightgray} 0.912 & \cellcolor{lightgray} 1.000 & -- \\
\bottomrule
\end{tabular}

}%
\end{table}

\begin{table}[ht]
\caption{Correlation coefficients between analyzed metrics, synthethic datasets. Values above the main diagonal are the Pearson correlation coefficient, while the\colorbox{lightgray}{values below}the main diagonal correspond to the Spearman's rank correlation.}
\label{tab:corr_mock}
\resizebox{\linewidth}{!}{%
\begin{tabular}{rcccccccccc}
\toprule
 & L1 $\downarrow$ & L2 $\downarrow$ & PSNR $\uparrow$ & SSIM $\uparrow$ & LPIPS $\downarrow$ & HaarPSI $\uparrow$ & $k$-off $\uparrow$ & Dice $\uparrow$ & Cr. entr. $\downarrow$ & Focal $\downarrow$ \\
\midrule
L1 $\downarrow$       & -- & 0.967 & -0.976 & -0.946 & -0.770 & 0.854 & -0.293 & -0.397 & -0.092 & 0.016 \\
    L2 $\downarrow$       & \cellcolor{lightgray} 0.954 & -- & -0.998 & -0.914 & -0.638 & 0.779 & -0.298 & -0.514 & -0.173 & -0.064 \\
    PSNR $\uparrow$       & \cellcolor{lightgray} -0.954 & \cellcolor{lightgray} -0.998 & -- & 0.930 & 0.664 & -0.793 & 0.253 & 0.421 & 0.216 & 0.132 \\
    SSIM $\uparrow$       & \cellcolor{lightgray} -0.884 & \cellcolor{lightgray} -0.868 & \cellcolor{lightgray} 0.868 & -- & 0.774 & -0.874 & 0.522 & 0.565 & -0.128 & -0.299 \\
    LPIPS $\downarrow$    & \cellcolor{lightgray} -0.417 & \cellcolor{lightgray} -0.348 & \cellcolor{lightgray} 0.365 & \cellcolor{lightgray} 0.356 & -- & -0.874 & -0.373 & -0.719 & 0.113 & 0.411 \\
    HaarPSI $\uparrow$    & \cellcolor{lightgray} 0.341 & \cellcolor{lightgray} 0.272 & \cellcolor{lightgray} -0.278 & \cellcolor{lightgray} -0.279 & \cellcolor{lightgray} -0.759 & -- & -0.091 & 0.242 & -0.088 & -0.188 \\
    $k$-off $\downarrow$  & \cellcolor{lightgray} -0.332 & \cellcolor{lightgray} -0.303 & \cellcolor{lightgray} 0.285 & \cellcolor{lightgray} 0.619 & \cellcolor{lightgray} -0.443 & \cellcolor{lightgray} -0.017 & -- & 0.477 & -0.080 & -0.237 \\
    Dice $\uparrow$       & \cellcolor{lightgray} -0.469 & \cellcolor{lightgray} -0.495 & \cellcolor{lightgray} 0.462 & \cellcolor{lightgray} 0.547 & \cellcolor{lightgray} -0.702 & \cellcolor{lightgray} 0.443 & \cellcolor{lightgray} 0.554 & -- & 0.259 & -0.123 \\
    Cr. entr. $\downarrow$& \cellcolor{lightgray} -0.134 & \cellcolor{lightgray} -0.204 & \cellcolor{lightgray} 0.252 & \cellcolor{lightgray} -0.132 & \cellcolor{lightgray} 0.126 & \cellcolor{lightgray} -0.087 & \cellcolor{lightgray} -0.096 & \cellcolor{lightgray} 0.226 & -- & 0.914 \\
    Focal $\downarrow$    & \cellcolor{lightgray} -0.012 & \cellcolor{lightgray} -0.092 & \cellcolor{lightgray} 0.155 & \cellcolor{lightgray} -0.232 & \cellcolor{lightgray} 0.226 & \cellcolor{lightgray} -0.066 & \cellcolor{lightgray} -0.289 & \cellcolor{lightgray} 0.023 & \cellcolor{lightgray} 0.954 & -- \\
\bottomrule
\end{tabular}

}%
\end{table}

\begin{table}[ht]
\caption{Correlation coefficients between analyzed metrics, real datasets. Values above the main diagonal are the Pearson correlation coefficient, while the\colorbox{lightgray}{values below}the main diagonal correspond to the Spearman's rank correlation.}
\label{tab:corr_real}
\resizebox{\linewidth}{!}{%
\begin{tabular}{rcccccccccc}
\toprule
 & L1 $\downarrow$ & L2 $\downarrow$ & PSNR $\uparrow$ & SSIM $\uparrow$ & LPIPS $\downarrow$ & HaarPSI $\uparrow$ & $k$-off $\uparrow$ & Dice $\uparrow$ & Cr. entr. $\downarrow$ & Focal $\downarrow$ \\
\midrule
L1 $\downarrow$       & -- & 0.728 & -0.856 & 0.098 & 0.659 & 0.003 & -0.603 & 0.509 & -0.602 & -0.587 \\
L2 $\downarrow$       & \cellcolor{lightgray} 0.605 & -- & -0.774 & 0.260 & 0.614 & -0.196 & -0.398 & 0.510 & -0.584 & -0.577 \\
PSNR $\uparrow$       &\cellcolor{lightgray} -0.950 &\cellcolor{lightgray} -0.568 & -- & -0.189 & -0.796 & 0.257 & 0.546 & -0.524 & 0.716 & 0.703 \\
SSIM $\uparrow$       &\cellcolor{lightgray} -0.373 &\cellcolor{lightgray} -0.025 &\cellcolor{lightgray} 0.428 & -- & 0.488 & 0.505 & -0.428 & 0.879 & -0.599 & -0.614 \\
LPIPS $\downarrow$    &\cellcolor{lightgray} 0.416 &\cellcolor{lightgray} 0.081 &\cellcolor{lightgray} -0.464 &\cellcolor{lightgray} -0.149 & -- & 0.088 & -0.586 & 0.832 & -0.788 & -0.781 \\
HaarPSI $\uparrow$    &\cellcolor{lightgray} -0.021 &\cellcolor{lightgray} -0.278 &\cellcolor{lightgray} 0.082 &\cellcolor{lightgray} 0.154 &\cellcolor{lightgray} -0.211 & -- & -0.052 & 0.459 & 0.003 & -0.010 \\
$k$-off $\downarrow$  &\cellcolor{lightgray} -0.354 &\cellcolor{lightgray} -0.238 &\cellcolor{lightgray} 0.215 &\cellcolor{lightgray} -0.314 &\cellcolor{lightgray} -0.342 &\cellcolor{lightgray} 0.013 & -- & -0.573 & 0.492 & 0.478 \\
Dice $\uparrow$       &\cellcolor{lightgray} 0.064 &\cellcolor{lightgray} 0.099 &\cellcolor{lightgray} 0.025 &\cellcolor{lightgray} 0.542 &\cellcolor{lightgray} 0.191 &\cellcolor{lightgray} 0.439 &\cellcolor{lightgray} -0.439 & -- & -0.721 & -0.725 \\
Cr. entr. $\downarrow$&\cellcolor{lightgray} 0.007 &\cellcolor{lightgray} -0.039 &\cellcolor{lightgray} 0.016 &\cellcolor{lightgray} -0.532 &\cellcolor{lightgray} -0.223 &\cellcolor{lightgray} -0.297 &\cellcolor{lightgray} 0.237 &\cellcolor{lightgray} -0.737 & -- & 0.999 \\
Focal $\downarrow$    &\cellcolor{lightgray} 0.033 &\cellcolor{lightgray} -0.062 &\cellcolor{lightgray} -0.016 &\cellcolor{lightgray} -0.568 &\cellcolor{lightgray} -0.190 &\cellcolor{lightgray} -0.296 &\cellcolor{lightgray} 0.236 &\cellcolor{lightgray} -0.747 &\cellcolor{lightgray} 0.994 & -- \\
\bottomrule
\end{tabular}

}%
\end{table}

\section{Supplementary 
training details}
\label{ap:training}

Models were implemented in PyTorch, and trained on compute nodes containing the NVIDIA A100 GPU.
The optimizer used was Adam, with a learning rate $lr=1e^{-4},  \beta_1=0.9, \beta_2 = 0.999$.
Stochastic gradient descent also shows satisfactory results. 
Different batch sizes were tried in preliminary training runs, from $32$ to $256$ patches per batch.
Perceptually better results were observed with $32$ samples per batch, which was therefore chosen for training.

After empirically fine-tuning, some models were trained with different variance thresholds $v$ and for more epochs.
Those are R2 on LPIPS ($v=20$); 
R3, R5 and R8 on LPIPS ($v=0$, 30 epochs);
R4 on L2 and cross entropy ($v=10$, 100 epochs); and all models for synthetic data on regression tasks ($v=0$, 30 epochs).

\section{Similarity measures of color quantized wafers through $k$-means clustering}

\begin{figure}
    \centering
    \includegraphics[width=0.95\linewidth]{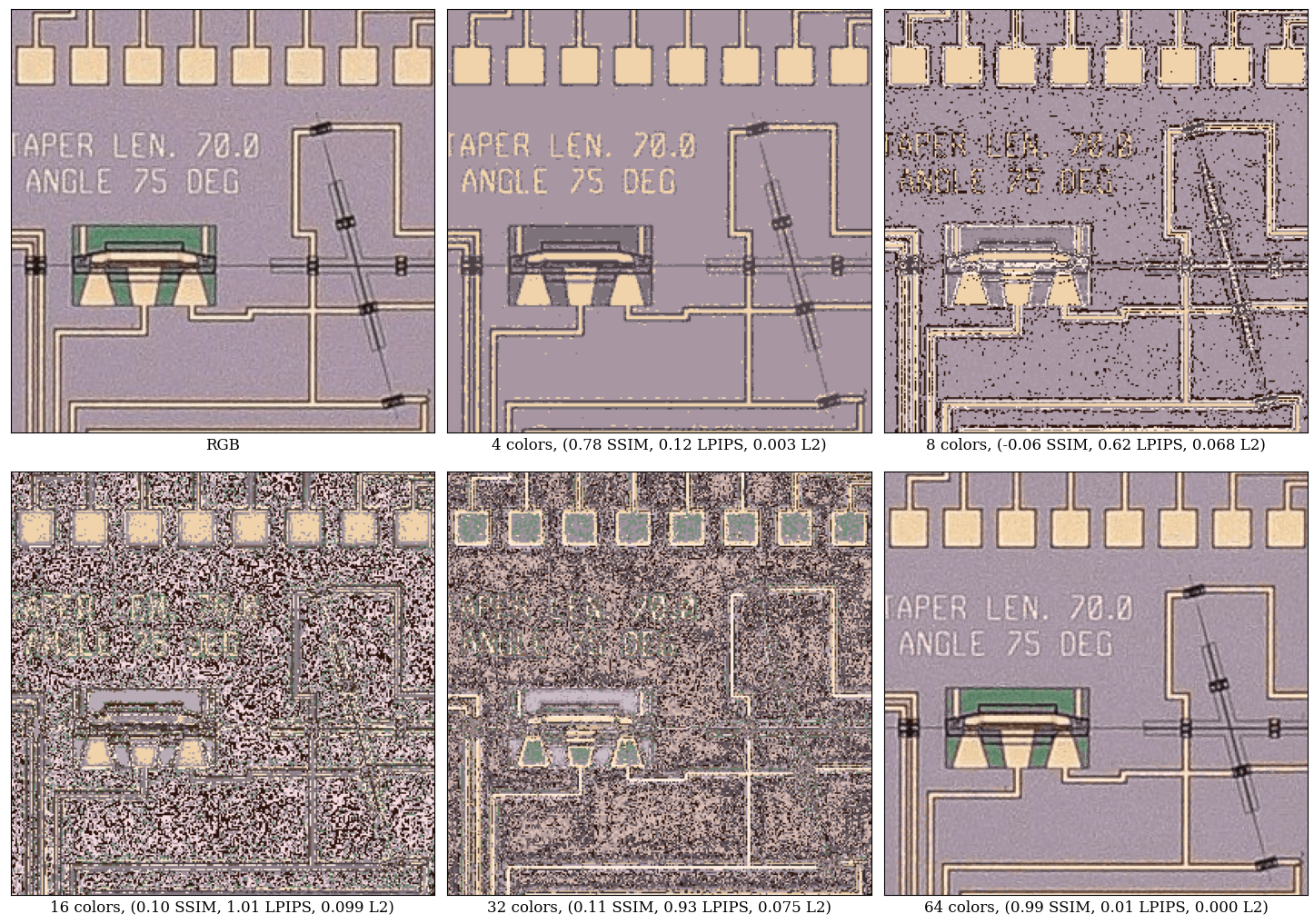}
    \caption{RGB reconstructions of quantized images for several amounts of colors --- i.e. different $k$ centroids in the $k$-means clustering method. The improvement in quality for $k=64$ is clear.}
    \label{fig:k64}
\end{figure}

The quantized color palette was obtained by $k$-means clustering of pixel colors of a random subsample of the original photographs.
The amount of centroids $k=64$ was chosen empirically and shows significant similarity to the original image.
It can be observed in Figure~\ref{fig:k64} how reconstructions for different $k$ appear.

\section{Dataset photograph samples}

\begin{figure}
    \centering
    \includegraphics[width=0.75\linewidth]{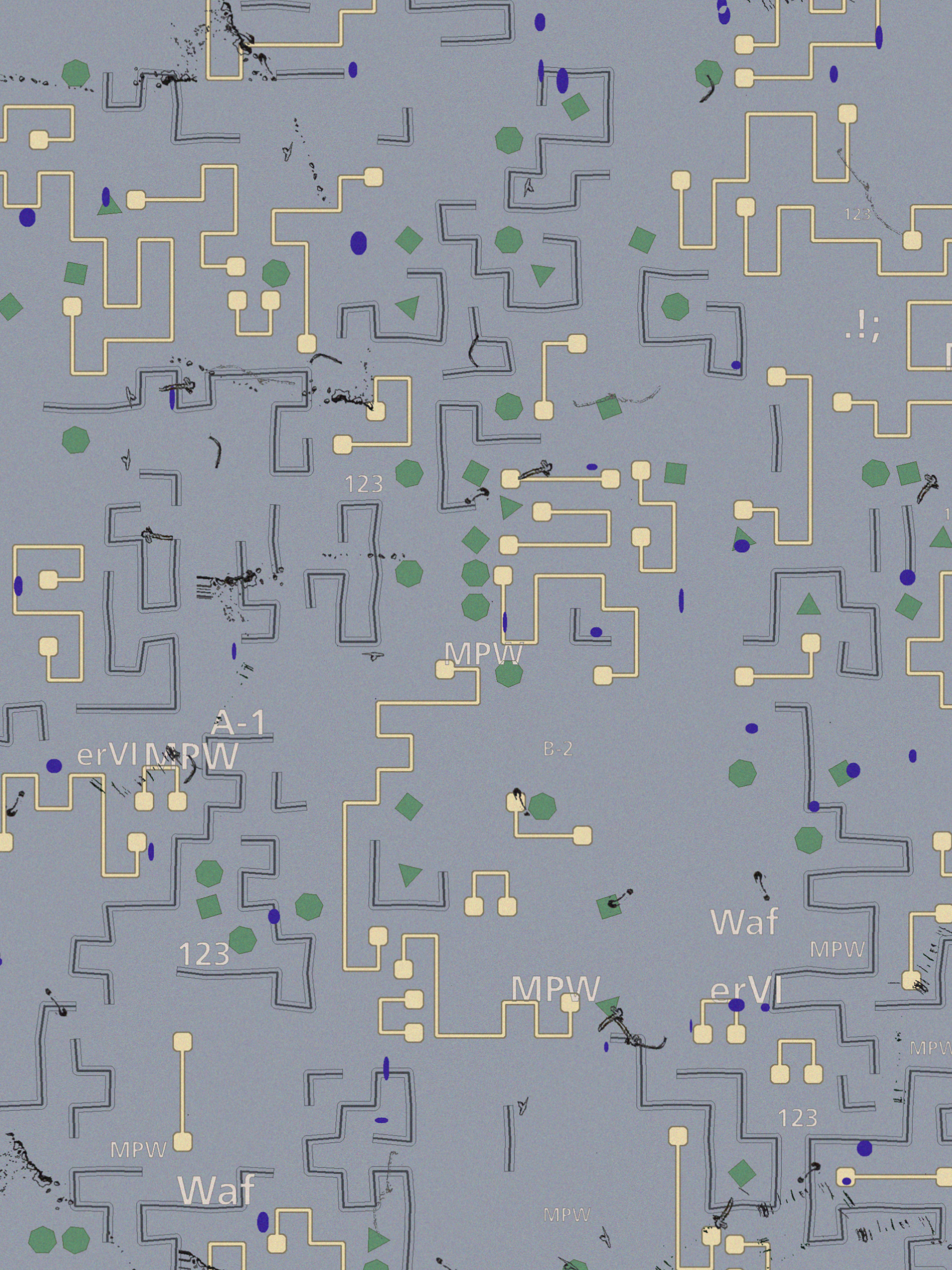}
    \caption{A 2000$\times$2667 pixel sample of S1, a synthetic dataset used in this work.}
    \label{fig:s1}
\end{figure}

\begin{figure}
    \centering
    \includegraphics[width=0.75\linewidth]{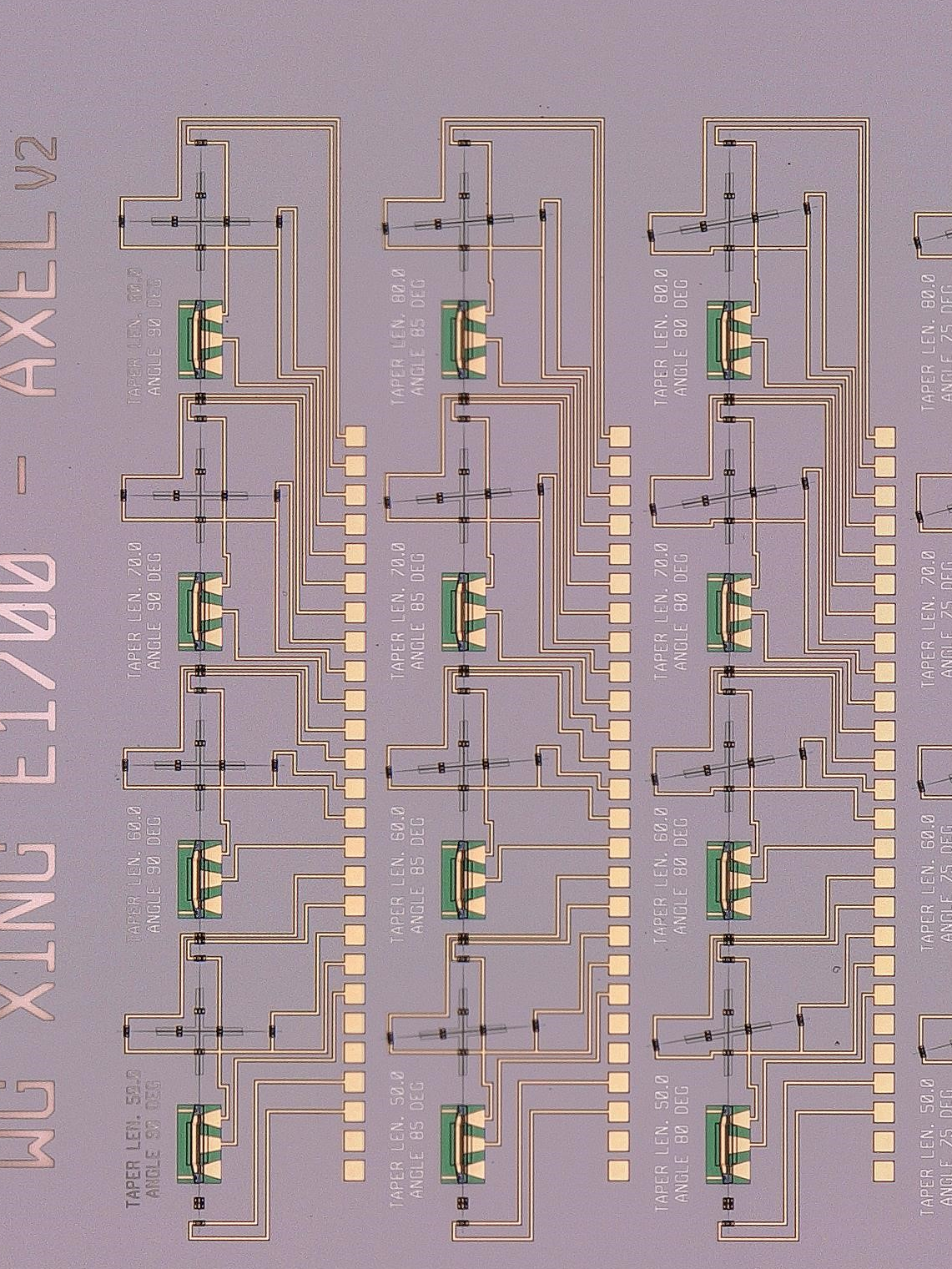}
    \caption[Real wafer sample]{A 1141$\times$1521 pixel sample of R5, a stitched photograph of a real wafer.}
    \label{fig:r5}
\end{figure}

We show a sample of a synthetic dataset (see Figure~\ref{fig:s1}) and of a real wafer (in Figure~\ref{fig:r5}).


\end{document}